\documentclass[twoside,11pt]{article}

\usepackage{jmlr2e}
\usepackage{graphicx}
\usepackage{amssymb}
\usepackage{subcaption}
\graphicspath{{Pics/}}
\usepackage{soul}
\usepackage{xcolor}
\usepackage{bm}
\usepackage{amsmath}
\usepackage[linesnumbered,ruled,vlined]{algorithm2e}
\usepackage{hyperref}

\DeclareMathOperator*{\E}{\mathbb{E}}

\newtheorem{definition1}{Definition}[section]

\ShortHeadings{GOLS for a variety of stochastic training algorithms}{Kafka and Wilke}
\firstpageno{1}

\begin{document}

\title{Gradient-only line searches to automatically determine learning rates for a variety of stochastic training algorithms}

\author{\name Dominic Kafka \email dominic.kafka@gmail.com\\
       \addr Centre for Asset and Integrity Management (C-AIM),
       Department of Mechanical and Aeronautical Engineering,
       University of Pretoria, Pretoria, South Africa
       \AND
       \name Daniel N. Wilke \email wilkedn@gmail.com \\
       \addr Centre for Asset and Integrity Management (C-AIM),
       Department of Mechanical and Aeronautical Engineering,
       University of Pretoria, Pretoria, South Africa}


\maketitle

\begin{abstract}

Gradient-only and probabilistic line searches have recently reintroduced the ability to adaptively determine learning rates in dynamic mini-batch sub-sampled neural network training. However, stochastic line searches are still in their infancy and thus call for an ongoing investigation. We study the application of the {\it Gradient-Only Line Search that is Inexact} (GOLS-I) to automatically determine the learning rate schedule for a selection of popular neural network training algorithms, including NAG, Adagrad, Adadelta, Adam and LBFGS, with numerous shallow, deep and convolutional neural network architectures trained on different datasets with various loss functions. We find that GOLS-I's learning rate schedules are competitive with manually tuned learning rates, over seven optimization algorithms, three types of neural network architecture, 23 datasets and two loss functions. We demonstrate that algorithms, which include dominant momentum characteristics, are not well suited to be used with GOLS-I. However, we find GOLS-I to be effective in automatically determining learning rate schedules over 15 orders of magnitude, for most popular neural network training algorithms, effectively removing the need to tune the sensitive hyperparameters of learning rate schedules in neural network training.


\end{abstract}

\begin{keywords}
 Artificial Neural Networks, Dynamic Mini-Batch Sub-Sampling, Search Directions, Learning Rates, Gradient-Only Line Search
\end{keywords}

\section{Introduction}


\begin{figure}[h!]
	\centering
	\includegraphics[width=0.6\linewidth]{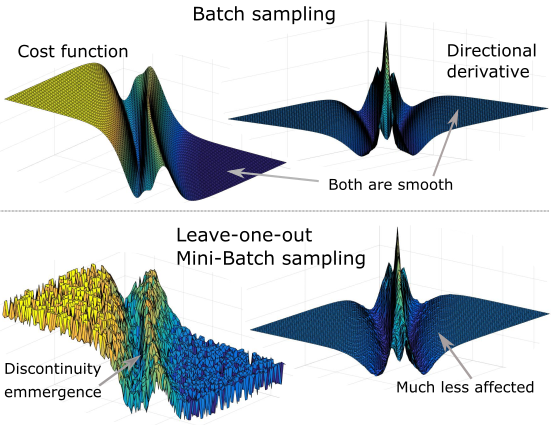}
	\caption{Dynamic mini-batch sub-sampling introduces discontinuities into the loss and gradients of a neural network. However, gradient information, depicted by the directional derivative along a unit direction of two weights, is less affected by discontinuities during resampling. Gradient-only line searches exploit this property to determine step sizes with the aid of the gradient-only optimality condition \citep{Wilke2013,Snyman2018,Kafka2019jogo,Kafka2019}.}
	\label{fig_intro}
\end{figure}

In pursuit of improving the training of neural networks, learning rate parameters are consistently some of the most sensitive hyperparameters in deep learning \citep{Bergstra2012}, and hence constitute an ongoing area of research \citep{Smith2015,Orabona2017,Wu2018}. The field of mathematical programming predominantly employs minimization line searches to determine step sizes (or learning rates) \citep{Arora2011}. However, ever since neural network training has moved from full-batch to mini-batch sub-sampled (MBSS) training to improve computational cost and training characteristics, line searches have fallen out of favour \citep{Wilson2003,Schraudolph2003,Schraudolph2006}. The prohibiting factor in successfully implementing minimization line searches has been the emergence of discontinuities in the loss functions and gradients, that is due to continuous resampling of mini-batches, see Figure~\ref{fig_intro}. Subsequently, the primary research focus in stochastic minimization has been on various {\it a priori} selected step sizes \citep{Schraudolph1999,Boyd2003,Smith2015}.

Recently, line searches were reintroduced to optimize dynamic MBSS loss functions in neural network training in the form of the probabilistic line search by \cite{Mahsereci2017a}.  This method deals with the discontinuities in function value and gradients by constructing Gaussian process surrogates along search directions. The minimum of the surrogate can subsequently be accurately determined. Alternatively, {\it gradient-only} line search methods \citep{Kafka2019jogo} employ concepts based on gradient-only optimization \citep{Wilke2013,Snyman2018} to determine step sizes. This is achieved by locating Stochastic Non-Negative Associated Gradient Projection Points (SNN-GPPs), which can be distinct from function minimizers in discontinuous functions. An SNN-GPP manifests as a sign change in directional derivative from negative to positive along a descent and search direction. This means, that the SNN-GPP definition includes second-order information, allowing the line search to avoid maxima and inflection points, searching only for minima that are present in the derivative, thereby reducing the number of candidate solutions. The use of the gradient-only paradigm has a double benefit: 1) Gradient information is less affected by discontinuities present in dynamic MBSS loss functions (see Figure~\ref{fig_intro}) and 2) the spatial location of SNN-GPPs remains bounded. When implemented, these factors have resulted in a gradient-only line search formulation, which has been shown to outperform probabilistic line searches in their current form \citep{Kafka2019}.

In this paper we investigate the suitability of using the {\it Gradient-Only Line Search that is Inexact} (GOLS-I), see Appendix \ref{sec_golsi}, to determine the learning rate parameters for a collection of seven popular neural network training algorithms on three types of neural network architectures, two loss functions and 23 datasets. This work fits into the broader context of improving optimization performance and efficiency in neural network training.


\section{Connections: Approaches in advancing neural network training}

Neural network training requires an optimization problem to be solved. Two fundamental components to any optimization problem are the formulation of the problem, with the resulting optimization landscape (called the loss function); and the behaviour of the optimizer selected to traverse the given landscape in search of an optimum (also referred to as the training algorithm). We briefly review both of these aspects in the context of neural network training.

\subsection{Modifying loss function landscapes}

Significant research has been directed towards theoretical characterization of the loss functions in neural network architectures. However, admittedly there is still a gap between theoretical basis and practical experiences in training. Hence there has been a movement to recast problems in manners that lessen the void between practice and more rigorously understood optimization theory \citep{Kawaguchi2016}.

Due to the non-convex nature of neural network loss functions, see Figure \ref{fig_intro}, many optimizers have difficulty finding optima. Therefore, there have been significant attempts to improve the characteristics of the loss function by, 
\begin{itemize}
	\item Loss function scaling for a given model \citep{Ioffe2017,Salimans2016},
	\item Regularization of a network model for convexity \citep{Bishop2006}, and
	\item Hyperparameter optimization to modify the network  \citep{Bergstra2011}.
\end{itemize} 
 
If length scales of the loss function in different dimensions are similar, the problem may become less sensitive to the training algorithm and learning rate selected. Methods such as both Batch Norm \citep{Ioffe2017} and Weight Norm \citep{Salimans2016} can be used to this effect. Batch Norm is explicitly   claimed to be effective in correcting the scaling of deep networks, where the exploding or diminishing gradient problems become prevalent in deeper layers, causing vastly differing curvatures along different directions.

A standard method of introducing convexity into the loss function landscape is by adding a regularization term such as a quadratic L2-normalization term to the network loss \citep{Bishop2006}. Alternatively, work done by \cite{Li2017} shows how architectural decisions, specifically the introduction of skip connections, in a neural network, affect the convexity of the loss function. Interesting investigative work done by \cite{Goodfellow2015} shows that networks, which perform well when used with stochastic gradient descent (SGD) often result in convex training paths. However, the authors do show that encountering narrow, flat ravines or flat planes is a possibility. Given this highly non-linear nature of loss functions, it is unlikely that a single fixed learning rate would perform well over all iterations.
\cite{Goodfellow2015} also demonstrated that another contributing factor to the performance of the algorithm was the directions generated for a given update step. If the variance in the generated directions is high, SGD follows trajectories which make little progress towards a minimum. This can be especially pronounced when the optimizer is traversing a flat plane with a small learning rate. These considerations are consistent with the observations by \cite{Smith2017}, which suggests finding better minima by ramping up the mini-batch size, thereby reducing gradient variance, as opposed to decreasing the learning rate.

Global hyperparameter optimization strategies consider the various parameters involved in problem formulation (network architecture, activation function, loss function etc.) and optimizer (update rule, learning rates etc.). Depending on the problem, there might be combinations of architecture and algorithm parameters that are more effective than others. Hyperparameter optimization approaches seek to isolate such combinations by changing both the loss landscape and optimization parameters simultaneously. Such methods include grid-searches, random search and Bayesian optimization \citep{Bergstra2011,Bergstra2012,Snoek2012}. However, they are generally expensive and require a large number of trial runs to evaluate the fitness of different parameter configurations. Although these methods are used to configure such a wide range of parameters, work done by \citet{Bergstra2011,Bergstra2012} show that the learning rate parameters are consistently the most sensitive across different problems. It is therefore of interest to determine the learning rate parameters automatically, making the remaining hyperparameter space smaller and easier to optimize.

\subsection{Improving optimizers}

In the interest of constructing effective optimizers for traversing the complex landscapes of neural network loss functions, research has been focussed on 
\begin{itemize}
	\item a priori selected step sizes \citep{Darken1990}, and
	\item adaptively computed step sizes \citep{Darken1990}.
\end{itemize}

An alternative to line searches is the use of {\it a priori} determined fixed learning rates and learning rate schedules, forming part of sub-gradient methods \citep{Boyd2003}. However, early work showed that fixed learning rates (step sizes) that are too small result in inefficient use of computational resources, while parameters that are too large inhibit optimization algorithms from converging to solutions with sufficient accuracy \citep{Moreira1995}. Hence, a more accessible approach in the deep learning community is to use learning rate schedules \citep{Darken1990,Moreira1995,Senior2013,Vaswani2017,Denkowski2017}, which change the learning rate according to a predetermined rule. Some schedules focus on incorporating oscillatory behaviour such as learning rate cycling \citep{Smith2015}, which ramps the learning rate up and down within a given range. A similar approach is the use of warm restarts \citep{Loshchilov2016}. Both these methods periodically vary their learning rates with regards to a specified function and schedule. This is inspired by allowing the optimization algorithm to balance exploration (high learning rate) and exploitation (small learning rate) to escape local minima or saddle points, and increase the odds of finding good minima. However, in general, the primary disadvantage of learning rate schedules is that both the functional form of the schedule, as well as its hyper-parameters, are problem dependent, but determined {\it a priori} to training, making generalized application of learning rate schedules difficult. That being said, popular optimizer improvements have included incorporating learning schedules directly into steepest descent algorithms \citep{Duchi2011,Zeiler2012,Kingma2015,Zheng2017,Wu2018}. However, in most cases, a learning rate magnitude parameter remains.

Another area of learning rate research is adaptive step size methods, which change parameters based on loss function information presented to the algorithm during training. 
This category includes line searches, which have been implemented in training machine learning problems in the field of adaptive sub-sampling. Here the emphasis lies on governing qualities of the training algorithm's descent direction through determining the mini-batch size. Once selected, mini-batches remain unchanged throughout an iteration \citep{Martens2010,Friedlander2011,Byrd2011,Byrd2012,Bollapragada2017,Kungurtsev2018,Paquette2018,Bergou2018,Mutschler2019}. However, \cite{Bottou2010} argues, that continually changing mini-batches can benefit training, by exposing the algorithm to increased amounts of information.

Consequently, alternative adaptive step size strategies have been developed, specifically to operate on discontinuous loss functions. Examples of such methods include the incorporation of statistical concepts into existing methods, such as the adaptation of coin betting theory to determine the learning rate \citep{Orabona2017} and a statistically motivated adaptation of an annealed learning rate \citep{Pouyanfar2017}. Interest also returned to the well established gradient descent with momentum algorithm \citep{Zhang2017}. The work by \cite{Zhang2017} demonstrated that by adaptively fine-tuning the momentum term, the performance of gradient descent with momentum could be competitive with state of the art algorithms while obtaining better generalization.  A particularly interesting albeit involved approach has been to use deep learning methods themselves to determine the learning rate during training \citep{Xu2017,Bello2017,Andrychowicz2016}. One approach has been to use a controller recurrent neural network to learn the appropriate update rule for the main network by assembling ingredients from a database of existing update rules \citep{Bello2017}. Another approach employs reinforcement learning by using the combination of an actor and critic network to learn and determine the learning rate schedule for the main network. This particular method trains 3 networks simultaneously \citep{Xu2017}. Such approaches can become computationally expensive, while trying to fill the void left by the absence of mainstream adoption of line searches.




\section{Our contribution}

Typical learning rate schedules generate values that exponentially decay from $0.1$ to $10^{-6}$ \citep{Senior2013}, while the magnitudes of the aforementioned cyclical learning rate schedules vary over 3 or 4 orders of magnitude \citep{Smith2015,Loshchilov2016}. Such schedules can have up to 6 different parameters which need to be set. The {\it Gradient-Only Line Search that is Inexact} (GOLS-I) \citep{Kafka2019jogo} is able to determine step sizes within a maximum range of 15 orders of magnitude and has no tunable hyperparameters. This allows the line search to traverse various loss function features, such as flat planes as well as steep slopes.

\begin{figure}[h!]
	\centering
	\includegraphics[width=0.7\linewidth]{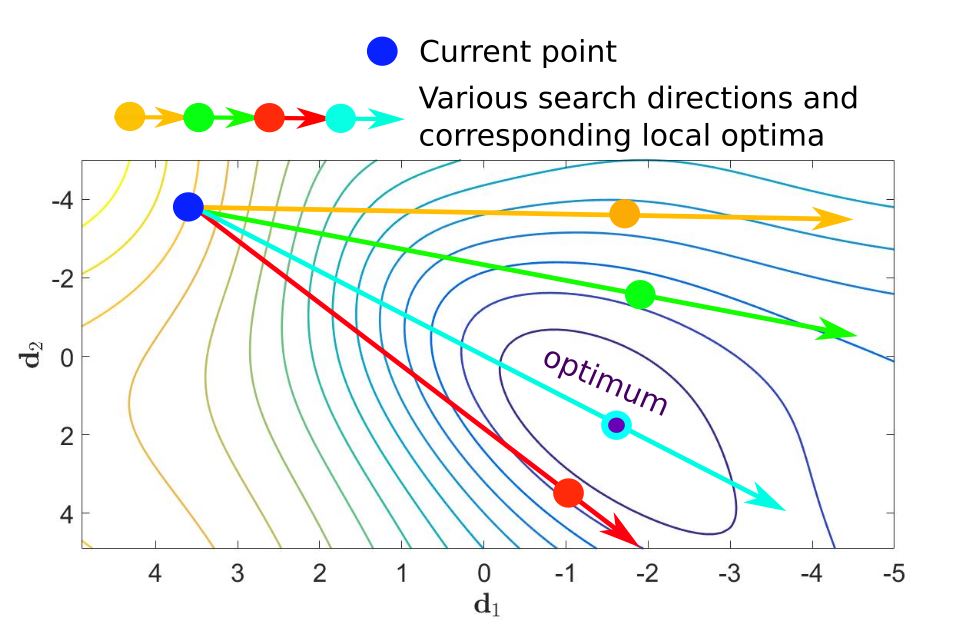}
	\caption{Not all directions are created equal. Different assumptions and paradigms result in a plethora of optimization strategies with varying search direction trajectories. Step sizes along these directions can be determined using line searches.}
	\label{fig_intro_dirs}
\end{figure}

We recast various popular training algorithms, namely (Stochastic) Gradient Descent (SGD) \citep{Robbins1951}, SGD with momentum (SGDM) \citep{Rumelhart1988}, Nesterov's Accelerated Gradient (NAG) \citep{Nesterov1983}, Adagrad \citep{Duchi2011}, Adadelta \citep{Zeiler2012} and Adam \citep{Kingma2015} to be compatible with line searches. Subsequently, we employ GOLS-I to determine the learning rates for these algorithms to explore whether the line search can adapt step sizes according to the characteristics of a given algorithm. Every algorithm contributes a search direction, for which GOLS-I is then tasked with determining a step size. This presents two questions: 1) Is GOLS-I able to adapt step sizes to different algorithms during training in the context of mini-batch sub-sampled loss functions? And 2) Which directions perform best with a line search, see Figure~\ref{fig_intro_dirs}. Arguably, none of the selected algorithms (apart from perhaps SGD as a derivative of gradient descent) were designed to be used with a line search, rendering the answer to the latter question non-self-evident.

In this study, we first explore the interaction between GOLS-I and the various algorithms by training 22 different foundational classification problems with both shallow and deep network architectures. Subsequently, we implement GOLS-I in PyTorch and apply it to the CIFAR10 dataset with the ResNet18 architecture to demonstrate how GOLS-I interacts with different algorithms on a larger problem.

\section{A summary of gradient-only line search concepts}
\label{sec_math}

Deep learning loss functions are predominantly formulated as
\begin{eqnarray}
\mathcal{L}(\boldsymbol{x}) = \frac{1}{M} \sum_{b=1}^{M} \ell (\boldsymbol{x};\;\boldsymbol{t}_b),
\label{eq:loss}
\end{eqnarray}
where $\boldsymbol{x}\in \mathcal{R}^p$ is a vector of network weights, $\{\boldsymbol{t}_1,\dots,\boldsymbol{t}_M\}$ is a training dataset with $M$ samples, and $\ell(\boldsymbol{x};\;\boldsymbol{t})$ is the loss formulation used to evaluate the performance of
weights $\boldsymbol{x}$ with regards to training dataset sample $\boldsymbol{t}$. Backpropagation \citep{Werbos1994} is then used to compute the exact gradient w.r.t.  $\boldsymbol{x}$ such that: 
\begin{eqnarray}
\nabla\mathcal{L}(\boldsymbol{x}) = \frac{1}{M} \sum_{b=1}^{M} \nabla\ell (\boldsymbol{x};\;\boldsymbol{t}_b).
\label{eq:lossgrad}
\end{eqnarray}

\begin{figure}[h!]
	\centering
	\includegraphics[width=0.8\linewidth]{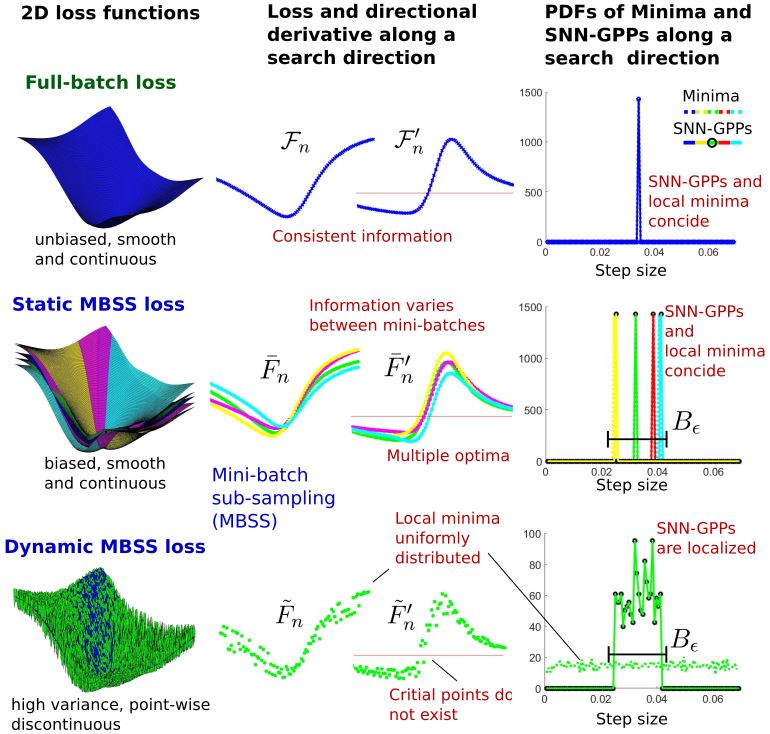}
	\caption{Summary of loss surfaces, static and dynamic mini-batch sub-sampling (MBSS) and the performance of minima vs SNN-GPPs.}
	\label{fig_intro_summary}
\end{figure}

In the case where the full batch of training data is used to evaluate $\mathcal{L}(\boldsymbol{x})$ and $\nabla\mathcal{L}(\boldsymbol{x})$, the resulting smoothness and continuity are only dictated by the smoothness and continuity properties of the activation functions. Consider Figures~\ref{fig_intro} and \ref{fig_intro_summary}, where we construct loss functions using the classic Iris \citep{Fisher1936} dataset with a simple feed-forward neural network with 10 hidden nodes.

A line search can then be conducted along a search direction $\boldsymbol{d}_n $ within the loss function, by constructing one-dimensional function $\mathcal{F}_n$ as a function of step size $\alpha$ as follows:
\begin{equation}
\mathcal{F}_n(\alpha) = f(\boldsymbol{x}_n(\alpha)) = \mathcal{L}(\boldsymbol{x}_n + \alpha \boldsymbol{d}_n),
\label{eq_linesearch}
\end{equation}
with corresponding directional derivative
\begin{equation}
\mathcal{F}'_n = \frac{d F_n(\alpha)}{d \alpha} = \boldsymbol{d}_n^T \cdot \nabla \mathcal{L}(\boldsymbol{x}_{n} + \alpha \boldsymbol{d}_n ), 
\label{eq_lineg}
\end{equation}
where $n$ denotes the iteration of a particular algorithm, currently located at $\boldsymbol{x}_n$. $F_n$ is a univariate search of the multi-dimensional loss function along the search direction $\boldsymbol{d}_n$, where $\mathcal{F}_n$ denotes a specific form of univariate search, where the loss function evaluated is $\mathcal{L}(\boldsymbol{x})$. An example is shown on the top of the middle column of Figure~\ref{fig_intro_summary}. A line search then seeks to find the optimum along the search direction of the loss function, see Figure~\ref{fig_intro_dirs}, which may require a number of function evaluations. We therefore denote step size as $\alpha_{n,I_n}$, where $n$ remains the iteration number, and $I_n$ is the number of function evaluations required to determine the step size for iteration $n$. This also generalizes to fixed-step sizes (learning rates), where $\alpha_{n,I_n}$ is selected {\it a priori} and $I_n=1$.

Minimization line searches have been shown to be effective in smooth, continuous functions \citep{Arora2011}. However, modern datasets, coupled with large neural network architectures \citep{Krizhevsky2012} exceed the available memory resources of computational nodes or graphical processing units (GPUs). Consequently, mini-batch sub-sampling (MBSS) is implemented, where only a fraction of the available training data $\mathcal{B} \subset \{1,\dots,M\}$ of size $|\mathcal{B}| \ll M$ is used to construct an approximate loss:

\begin{equation}
L(\boldsymbol{x}) = \frac{1}{|\mathcal{B}|} \sum_{b\in \mathcal{B}} \ell (\boldsymbol{x};\;\boldsymbol{t}_b),
\label{eq:lossgradbatch}
\end{equation}
and corresponding approximate gradient
\begin{equation}
\boldsymbol{g}(\boldsymbol{x}) = \frac{1}{|\mathcal{B}|} \sum_{b\in \mathcal{B}} \nabla\ell (\boldsymbol{x};\;\boldsymbol{t}_b).
\label{eq:g_lossgradbatch}
\end{equation}

These approximations then replace $\mathcal{L}(\boldsymbol{x})$ and $\nabla \mathcal{L}(\boldsymbol{x})$ for line searches in Equations (\ref{eq_linesearch}) and (\ref{eq_lineg}) respectively. Both approximations have respective expectations $\E [ L(\boldsymbol{x}) ] = \mathcal{L}(\boldsymbol{x})$ and $\E [ \boldsymbol{g}(\boldsymbol{x}) ] = \nabla\mathcal{L}(\boldsymbol{x})$ \citep{Tong2005}. However, individual evaluations of $L(\boldsymbol{x})$ and $\boldsymbol{g}(\boldsymbol{x})$ may vary significantly from the expectation, resulting in a sampling bias for each individual mini-batch, and variance between subsequent mini-batches.

In the field of adaptive sub-sampling methods, the emphasis lies on determining $|\mathcal{B}|$ or sample constituents of $\mathcal{B}$ that result in desired characteristics of $L(\boldsymbol{x})$. The aim might be to select $\mathcal{B}$, such that $L(\boldsymbol{x}) \approx \mathcal{L}(\boldsymbol{x})$, or to construct $L(\boldsymbol{x})$ such that it on average presents descent directions to the given algorithm \citep{Friedlander2011,Bollapragada2017}. Since much computational effort goes into selecting $\mathcal{B}$ for estimation $L(\boldsymbol{x})$, it is maintained over a minimum duration of a line search at iteration, $n$, in a given algorithm, before $\mathcal{B}$ is resampled \citep{Byrd2011,Byrd2012,Martens2010}. This approach is referred to as static MBSS, denoted as $\bar{L}(\boldsymbol{x})$, which presents smooth, but biased estimates to the optimization algorithm. A mini-batch sampled using static MBSS is denoted as $\mathcal{B}_n$ \citep{Kafka2019jogo}. We give a specific example in Figure~\ref{fig_intro_summary}, where we divide the training data into four mini-batches, the content of which remain constant, resulting in four different expressions of $\bar{L}(\boldsymbol{x})$ and $\bar{\boldsymbol{g}}(\boldsymbol{x})$. Corresponding line searches would be denoted $\bar{F}_n$ with derivative $\bar{F}'_n$. This approach allows minimization line searches to be used within given optimization algorithms. However, in our example, a line search will find one of four different minimizers, depending on which mini-batch is selected.

An alternative sampling strategy is dynamic MBSS, in which a new sub-sample of training data is selected at every evaluation, $i$, of ${L}(\boldsymbol{x})$, subsequently denoted as $\tilde{L}(\boldsymbol{x})$ and $\tilde{\boldsymbol{g}}(\boldsymbol{x})$ with line searches $\tilde{F}_n$ and $\tilde{F}'_n$. Since mini-batches are resampled at every function evaluation, $i$, they are correspondingly denoted as $\mathcal{B}_{n,i}$. In the example shown in Figure~\ref{fig_intro_summary} (last row), we uniformly sample one of the four mini-batches (previously constructed for the static MBSS example) for each function evaluation. The resulting loss function estimates are discontinuous due to continually changing mini-batches, which breaks the continuity of information between function evaluations. These discontinuities make encountering a critical point, $\tilde{F}'_n(\alpha^*) = 0$ \citep{Arora2011}, unlikely, even when $\mathcal{F}'_n(\alpha^*)=0$ for the full-batch loss. Additionally, minimization line searches often wrongly identify discontinuities as local minima \citep{Wilson2003,Schraudolph2003,Schraudolph2006}, motivating the aforementioned shift to subgradient methods \citep{Schraudolph1999,Boyd2003,Smith2015}. Implementing dynamic MBSS has also been referred to as approximate optimization \citep{Bottou2010}, having the benefit of exposing algorithms to larger amounts of information, due to continuous resampling of data. Line searches were first introduced into dynamic MBSS loss functions by \cite{Mahsereci2017a} and are subsequently receiving increased attention \citep{Mahsereci2017a,Wills2018,Wills2019,Kafka2019jogo,Kafka2019}.

The GOLS-I method introduced by \cite{Kafka2019jogo} employs the stochastic {\it gradient-only} equivalent to the optimality criterion in the form of the {\it Stochastic Non-Negative Associative Gradient Projection Point} (SNN-GPP) \citep{Kafka2019jogo}, which forms part of gradient-only optimization \citep{Wilke2013,Snyman2018}. The work done by \cite{Kafka2019jogo} presents proofs that GOLS-I converges within a bounded ball, $B_\epsilon$, where the ball bounds all possible SNN-GPPs of the surrounding neighbourhood. The SNN-GPP itself is defined as follows:

\begin{definition1}{SNN-GPP:}
	A stochastic non-negative associated gradient projection point (SNN-GPP) is defined as any point, $\boldsymbol{x}_{snngpp}$, for which there exists  $r_u > 0$ such that 
	\begin{equation}
	\nabla f(\boldsymbol{x}_{snngpp} + \lambda \boldsymbol{u} )\boldsymbol{u} \geq 0,\;\;  \forall\; \boldsymbol{u} \in \left\{\boldsymbol{y}\in\mathbb{R}^p \;|\; \| \boldsymbol{y} \|_2 = 1\right\},\;\;\forall \;\lambda \in (0,r_u]
	\end{equation}	
	with non-zero probability. \citep{Kafka2019jogo}
	\label{def_snngpp}
\end{definition1}

Along a descent direction, an SNN-GPP manifests as a directional derivative sign change from negative to positive. This definition makes allowance for the case where $\tilde{F}_n(\alpha^*) = 0$ may not exist at $\alpha^*$, while also generalizing to the case where $\mathcal{F}(\alpha^*) = 0$ or $\bar{F}(\alpha) = 0$. In the latter two continuous cases, the probability of encountering an SNN-GPP is 1. We numerically demonstrate this assertion by estimating the probability density functions of encountering minima and SNN-GPPs along a search direction in the third column of Figure~\ref{fig_intro_summary}. We conduct 100 line searches with 100 increments of fixed-step size $\alpha_{n,I_n} = 0.0007$ along the steepest descent direction for the Iris problem. We capture the frequency and location of minima and SNN-GPPs along the search direction and divide the frequency by the total number of occurrence to estimate the probability density function of minima and SNN-GPPs occurring. For continuous functions, all the probability of encountering a minimum or SNN-GPP is isolated to at $\alpha^* = \alpha_{snngpp}$. The same applies to static MBSS, with the only difference being that each mini-batch has an optimum at a different location. We call the spatial range of these separate mini-batch optima $B_\epsilon$, which essentially defines a 1-D ball.

In the case of dynamic MBSS, the sampling induced discontinuities cause local minima to occur uniformly across the sampled domain. Importantly, the same does not apply to SNN-GPPs, which remain bound within the range $B_\epsilon$. This bound depends on the variance of $\tilde{F}'_n(\alpha)$ across mini-batches sampled using dynamic MBSS. In our foundational example, the range of $B_\epsilon$ has defined edges, as there are only 4 mini-batches, with distinct locations of SNN-GPP's between the step sizes $\alpha \in [0.025,0.04]$. Outside of $B_\epsilon$, the probability of encountering a sign change is zero for the given search direction, regardless of the order in which the mini-batches are selected. However, within $B_\epsilon$ sign changes can occur at varying probabilities, depending on how mini-batches are selected. An SNN-GPP will only be encountered when a previous mini-batch results in a negative directional derivative, and the next results in a positive directional derivative. This can only occur within the range of the first and last mini-batch's sign change along a search direction.

\begin{figure}[h!]
	
	\centering
	\begin{subfigure}{.49\textwidth}
		\centering 
		\includegraphics[width=0.85\linewidth]{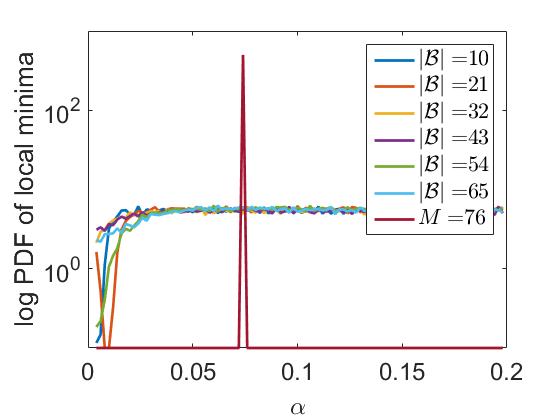}
		\caption{Function values}
		\label{fig_dists_d11_f}
	\end{subfigure}%
	\begin{subfigure}{.49\textwidth}
		\centering
		\includegraphics[width=0.85\linewidth]{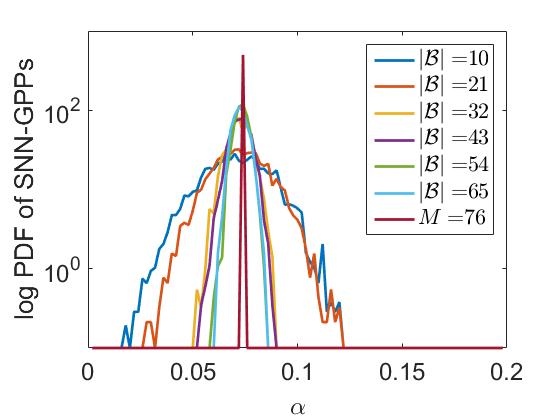}
		\caption{Directional derivatives}
		\label{fig_dists_d11_g}
	\end{subfigure}%
	\caption{Comparison between the estimated PDFs of (a) local minima and (b) SNN-GPPs along the steepest descent direction using uniformly sampled mini-batches, for the Iris dataset with mini-batch sizes ranging from $|\mathcal{B}_{n,i}| = 10$ to $M = 76$ (subscripts omitted for compactness). The x-axis indicates the step size $\alpha$ in 100 increments along $\boldsymbol{d} = -\nabla \mathcal{L}(\boldsymbol{x})$ from a fixed starting point. The y-axis shows the log of the probability of encountering a local minimum or SNN-GPP along the search direction. SNN-GPPs are more localized around the true optimum at $\approx 0.076$ with inaccuracy proportionate to sample size. Local minima are approximately uniformly distributed about the sampled domain for all sample sizes but $M$.}
	\label{fig_func_d11}
\end{figure}

However, unlike in our foundational example, most practical implementations of dynamic MBSS in neural network training uniformly sample the content of mini-batches, $\mathcal{B}_{n,i}$, directly from the training set. In Figure~\ref{fig_func_d11} we sample $\mathcal{B}_{n,i}$ with different batch sizes, ranging from 10 samples to the full selected training set of 76 samples. The plots shown are the result of conducting 500 runs over $\alpha_{n,I_n} \in \{0,0.002,0.004, \dots, 0.2\}$ and estimating the PDFs of locating local minima and SNN-GPPs along the steepest descent direction as obtained from the full-batch sampled loss, $\boldsymbol{d} = -\nabla \mathcal{L}(\boldsymbol{x}_n)$ at fixed starting point $\boldsymbol{x}_n$. However, here we show the distributions in the log domain for a clearer comparison. For line search methods, a tightly bound distribution is desired, as it indicates the location of an optimum with high reliability. Here, the distribution of SNN-GPPs approximates a Gaussian form, centred around the true optimum and with increasing variance as $|\mathcal{B}_{n,i}|$ is reduced. Recent work suggests that the variance in sampled gradients is representative of $\alpha$-distributions with $\alpha < 2$ \citep{Simsekli2019a}, implying that the tails of the distributions are longer than those of Gaussians. The tails of $\alpha$-distributions, like Gaussians, eventually tend towards 0, meaning that the likelihood of encountering an SNN-GPP diminishes far from the true optimum. However, this relationship does not hold for encountering local minima in dynamically MBSS losses, as shown in Figure~\ref{fig_func_d11}(a). The distribution of candidate minimizers in Figure~\ref{fig_func_d11}(a) approximates a uniform distribution over the sampled domain and is mainly independent of mini-batch size, which is consistent with our foundational problem in Figure~\ref{fig_intro_summary}. It is only for the full-batch loss functions, that the location of the local minimum is bounded.

This experiment demonstrates the benefit of searching for SNN-GPPs when implementing dynamic MBSS rather than function minimizers. The location of SNN-GPPs is bounded, meaning that an estimate of the location of the true, full-batch minimum can be obtained. GOLS-I employs an inexact line search approach, which interacts well with the approximate nature of SNN-GPPs, while also reducing the computational cost of the method compared to exact line searches \citep{Kafka2019jogo}. The availability of GOLS-I to determine step sizes for dynamic MBSS loss functions leads to the question as to how the method interacts with existing popular neural network training algorithms.

\section{Training algorithms and constructing search directions}

Recall the algorithms selected for this study, namely: (Stochastic) Gradient Descent (SGD)\citep{Morse1954}, SGD with momentum (SGDM) \citep{Rumelhart1988}, Nesterov's Accelerated Gradient (NAG) \citep{Nesterov1983}, Adagrad \citep{Duchi2011}, Adadelta \citep{Zeiler2012} and Adam \citep{Kingma2015}. 

These algorithms can be divided into two groups: 
\begin{enumerate}
	\item Coupled directions algorithm class: SGD, SGDM and NAG, in which learning rates operate only on the length scale of search directions. The ratios between components in the search direction remain fixed, while the learning rate modifies the magnitude of the search direction vector. And,
	\item Uncoupled directions algorithm class: Adagrad, Adadelta and Adam. These algorithms incorporate separate learning rate schedules along each component of the search direction, based on the respective component's historical gradient information. This results in search directions that vary significantly in both magnitude and orientation from historical gradient vectors.
\end{enumerate}

The coupled class encompasses SGD as well as alterations to its formulation to alleviate unwanted behaviour. The addition of momentum is an attempt to allow SGD the generation of mild ascent directions to overcome small local minima. It also alleviates the consecutively orthogonal direction behaviour of gradient descent \citep{Arora2011} that would also apply to SGD, instead of constructing directions that follow the contours of the loss function more closely. The NAG algorithm is a further improvement of SGD with momentum, incorporating a pre-emptive update step. 

In contrast, the uncoupled class of algorithms applies a learning rate schedule along each component of the weight vector, which is dependent on the history of gradient magnitudes in each respective dimension. These algorithms have shown to be effective in generating search directions for the non-convex loss functions of deep learning tasks. Each algorithm in this group proposes a different schedule. Adagrad proposes an inverse relationship of the step size to the sum of squared gradient components. The result of this is that as the sum increases, the step size decreases, which can be problematic during training runs of many iterations, as updates along dimensions with large gradient component magnitudes quickly tend to zero. Adadelta and Adam are two different formulations that seek to rectify this behaviour.  

Importantly, all of the mentioned algorithms (except Adadelta) still have a learning rate related hyperparameter that needs to be determined. To integrate line searches into these algorithms, consider the formulation of the standard update step:

\begin{equation}
\boldsymbol{x}_{n+1} = \boldsymbol{x}_n + \alpha_{n,I_n} \boldsymbol{u}_n,
\label{eq_update}
\end{equation} 
with the unit vector $\boldsymbol{u} = \frac{\boldsymbol{d}_n}{\|\boldsymbol{d}_n\|}$ dictating the unit search direction along which an update is to be made, while $\alpha_{n,I_n} \|\boldsymbol{u}_n\|$ determines the distance or step along the direction to be taken. This framework uncouples the direction of the update step, $\boldsymbol{u}_n$ from its magnitude, $\alpha_{n,I_n}$, which can be useful for interpretation of the problem in the context of a line search. However, it is the unscaled direction $\boldsymbol{d}_n$ that is given by most learning algorithms. Evaluating the norm thereof is unnecessary since a line search can compensate for the magnitude of $\boldsymbol{d}_n$ directly. As we are predominantly interested in determining the step size required for the given direction, we pose all given algorithms in the form:
\begin{equation}
\boldsymbol{x}_{n+1} = \boldsymbol{x}_n + \alpha_{n,I_n} \boldsymbol{d}_n.
\label{eq_updated}
\end{equation} 
Although some of the algorithms considered in this study were not proposed initially within this framework, we have taken the liberty to cast them into the form of Equation (\ref{eq_updated}), as given in Appendix \ref{sec_algs}. This allows us to determine $\alpha_{n,I_n}$ for the various algorithms using several line search methods \citep{Kafka2019jogo}. We augment the name of the training algorithm with the prefix "LS-" in Appendix~\ref{sec_algs} to accommodate the name of the line search. Therefore, as an example, implementing GOLS-I to determine $\alpha_{n,I_n}$ for LS-SGD is called "GOLS-I SGD". Note, that the original formulations of these algorithms can be recovered by using fixed-step sizes for $\alpha_{n,I_n}$. This concludes the ingredients required to train neural networks using gradient-only line searches with the search directions given by the selected algorithms.



\subsection{Available Software}

For the sake of transparency and reproducibility, our code is available at \url{https://github.com/gorglab/GOLS}. We have included accessible versions of the source code used for our investigations in the given repositories. Included are GPU compatible examples of other GOLS methods as developed in \cite{Kafka2019jogo}. These can also be coupled with any of the training algorithms considered in this study. The examples presented are self-contained, allowing for an environment well suited to exploring the characteristics of GOLS.

\section{Numerical studies}

We consider the 7 chosen training algorithms combined with 23 different datasets, and in total 3 different types of network architecture, namely single hidden layer feed-forward networks, deep feed-forward networks and the ResNet18 convolutional network architecture. We first consider performances of individual datasets and training algorithms, exploring the interaction between GOLS-I on a variety of problems. We compare GOLS-I to 3 fixed-step sizes ranging over 3 orders of magnitude. Importantly, the fixed-step sizes are manually chosen to demonstrate a range of training performances that are slow, efficient and unstable, respectively. Conversely, GOLS-I is implemented without any user intervention and performance is compared to that of manually chosen learning rates, while the computational cost of selecting appropriate fixed learning rates is omitted in the comparison. Thereafter we explore the average performances of GOLS-I over the pool of datasets for the various algorithms. 

Although single hidden layer neural networks have fallen out of favour over deep neural networks, they still offer significant research value to identify insight and isolate the essence concepts. Subsequently, we extend the investigation to deep networks of up to 6 hidden layers to demonstrate the robustness of GOLS-I in the context of deep learning. Finally, we implement GOLS-I for training the ResNet18 architecture with 4 Convolutional layers on the well known Cifar10 dataset.

\subsection{Single hidden layer feed-forward neural networks}
\label{sec_SHLNN}

We implement single hidden layer feed-forward networks with the mean squared error (MSE) loss \citep{Prechelt1994} to 22 classification datasets. The datasets span from 1936 to 2016 with sizes that vary from 150 to 70 000 observations. These are used to investigate the generality of combining GOLS-I with different training algorithms on datasets with diverse characteristics. The details of the chosen datasets and associated parameters with network architectures are summarized in Table \ref{tbl_datasets_c4}. The number of nodes in the hidden layer for every dataset was determined by a combination of a proposed guideline \citep{Yu1992} and a regression upper bound, discussed in Appendix \ref{app_heuristics}. 

The selected fixed-step sizes are split into small, medium and large respectively, each experimentally determined by conducting several training runs over all datasets. The magnitudes were chosen such that the medium fixed-step sizes would give the most competitive performance, while smaller steps would cause slow convergence, and larger steps could result in divergent behaviour. The range between the smallest and largest fixed-step sizes was fixed at 3 orders of magnitude to highlight the sensitivity of these parameters, see Table \ref{tbl_learning_rates}. However, the learning rates for the MNIST problem were more sensitive than the rest, such that the large fixed step was reduced to the point where the algorithms did not crash. We initialize GOLS-I with its minimum step size, $\alpha_{min}=1\cdot 10^{-8}$, to allow the line search automatically adapt the step size to the given problem. This avoids incorporating bias into the line search, which might suit a particular problem.

\begin{table}[h!]
	\centering
	\caption{Learning rate ranges as used for different optimization algorithms.}
	\scalebox{0.7}{
		\begin{tabular}{|c|c|c|c|}
			\hline  & \textbf{LS-SGD, LS-NAG} & \textbf{LS-SGDM,} \textbf{LS-Adadelta}  & \textbf{LS-Adagrad, LS-Adam} \\ 
			\hline \textbf{Small Fixed Step} & 1 & 0.1 & 0.01 \\ 
			\hline \textbf{Medium Fixed Step} & 10 & 1 & 0.1 \\ 
			\hline \textbf{Large Fixed Step} & 100 & 10 & 1 \\ 
			\hline 
	\end{tabular} }
	\label{tbl_learning_rates}
\end{table}

\begin{table}[h!]
	\centering
	\caption{Properties of the datasets considered for the investigation with the corresponding neural network architecture.}
	\scalebox{0.7}{
		\begin{tabular}{|c|c|c|c|c|c|c|}
			\hline  
			& \multicolumn{5}{|c|}{\textbf{Dataset properties}} & \multicolumn{1}{c|}{\textbf{ANN properties}}\\
			\hline
			\textbf{No.} & \textbf{Dataset name} & \textbf{Author} & \textbf{Observations,} $M$  & \textbf{Inputs,} $D$ & \textbf{Classes,} $K$ & \textbf{Hidden nodes,} $H$ \\ 
			\hline \textbf{1} & Cancer1 & \citet{Prechelt1994} & 699  & 9 & 2 & 8  \\ 
			\hline \textbf{2} & Card1 & \citet{Prechelt1994} & 690  & 51 & 2 & 5  \\ 
			\hline \textbf{3} & Diabetes1 & \citet{Prechelt1994} & 768  & 8 & 2 & 7 \\ 
			\hline \textbf{4} & Gene1 & \citet{Prechelt1994} & 3175 & 120 & 3 & 9 \\ 
			\hline \textbf{5} & Glass1 & \citet{Prechelt1994} & 214 & 9 & 6 & 5 \\ 
			\hline \textbf{6} & Heartc1 & \citet{Prechelt1994} & 920 & 35 & 2 & 3 \\ 
			\hline \textbf{7} & Horse1 & \citet{Prechelt1994} & 364 & 58 & 3 & 2 \\ 
			\hline \textbf{8} & Mushroom1 & \citet{Prechelt1994} & 8124 & 125 & 2 & 22 \\ 
			\hline \textbf{9} & Soybean1 & \citet{Prechelt1994} & 683 & 35 & 19 & 3 \\ 
			\hline \textbf{10} & Thyroid1 & \citet{Prechelt1994} & 7200 & 21 & 3 & 20 \\ 
			\hline \textbf{11} & Abalone & \citet{Nash1994} & 4177 & 8 & 29 & 7 \\ 
			\hline \textbf{12} & Iris & \citet{Fisher1936} & 150 & 4 & 3 & 3 \\ 
			\hline \textbf{13} & H.A.R. & \citet{Anguita2012} & 10299 & 561 & 6 & 7\\ 
			\hline \textbf{14} & Bankrupted Co. (yr. 1) & \citet{Zieba2016} & 7027 & 64 & 2 & 35\\ 
			\hline \textbf{15} & Defaulted Credit Cards & \citet{Yeh2009} & 30000 & 24 & 2 & 23\\ 
			\hline \textbf{16} & Forests & \citet{Johnson2012} & 523 & 27 & 4 & 6\\ 
			\hline \textbf{17} & FT Clave & \citet{Vurkac2011} & 10800 & 16 & 4 & 15\\ 
			\hline \textbf{18} & Sensorless Drive & \citet{Paschke2013} & 58509 & 48 & 11 & 47 \\ 
			\hline \textbf{19} & Wilt & \citet{Johnson2013a} & 4839 & 5 & 2 & 4 \\ 
			\hline \textbf{20} & Biodegradable Compounds & \citet{Mansouri2013} & 1054 & 41 & 2 & 8\\ 
			\hline \textbf{21} & Simulation Failure & \citet{Lucas2013} & 540 & 20 & 2 & 8\\ 
			\hline \textbf{22} & MNIST Handwriting & \citet{Lecun1998} & 70000 & 784 & 10 & 30\\ 
			\hline 
	\end{tabular}}
	\label{tbl_datasets_c4} 
\end{table}

Once established, no adjustments were made to algorithms between application to different datasets. All datasets were prepared according to well known guidelines set out by \citet{Prechelt1994} and divided into a training, test and validation dataset respectively by the ratios 2:1:1. The training set is used to optimize the model, while we consciously split the remaining data into validation and test datasets to show the generality of the training algorithms. Validation datasets are typically used to govern stopping criteria or tune other hyperparameters, while the test dataset is used to assess the utlimate fitness of the network. Therefore, a good correlation between validation and test dataset performances is desired, such that calibration on the validation dataset results in network performance that transfers well to the test dataset. Though we did not do any hyperparameter tuning, and our training runs were conducted by limiting the number of iterations, we include both validation and test dataset loss evaluations in our study to demonstrate that both validation and test datasets were large enough to represent generalization performance fairly. This avoids selecting biased "generalization" data subsets, as most of the considered datasets do not have predetermined training, validation and test dataset separations.  

During training, the mini-batch sizes were kept constant at $|\mathcal{B}_{n,i}| = 32$, where each batch was uniformly selected with a replacement for every evaluation of $\tilde{L}(\boldsymbol{x})$ and $\tilde{\boldsymbol{g}}(\boldsymbol{x})$ across all datasets. Training, validation and test dataset loss evaluations were averaged over 10 training runs of 3000 iterations each. Initial network configurations, $\boldsymbol{x}_0$, were randomly sampled from a uniform distribution with range $[-0.1, 0.1]$. These experiments were conducted using Matlab \citep{MatlabR2015b}.


\subsubsection{GOLS-I adapting to different datasets using LS-SGD}
\label{sec_golsData}

\begin{figure}[h]
	\begin{subfigure}{.33\textwidth}
		\includegraphics[width=0.95\linewidth]{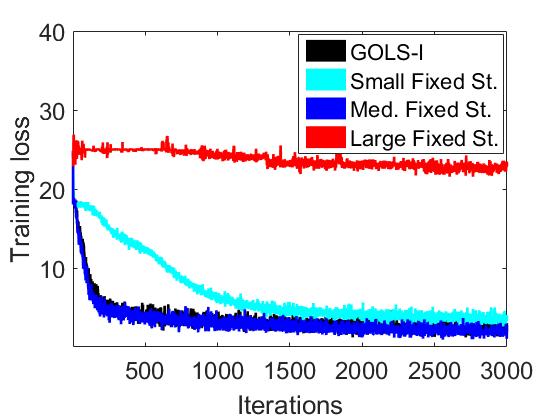}
		\caption{Forest dataset}
	\end{subfigure}%
	\begin{subfigure}{.33\textwidth}
		\includegraphics[width=0.95\linewidth]{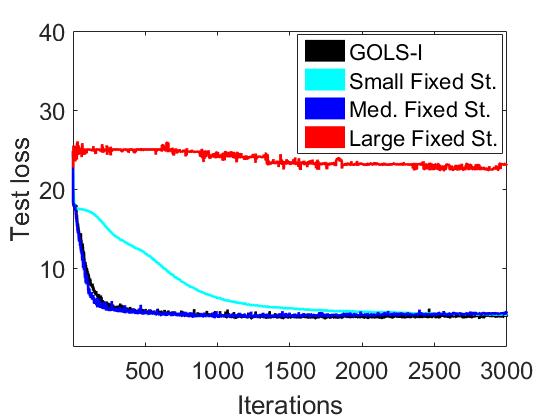}
		\caption{Forest dataset}
	\end{subfigure}%
	\begin{subfigure}{.33\textwidth}
		\includegraphics[width=0.95\linewidth]{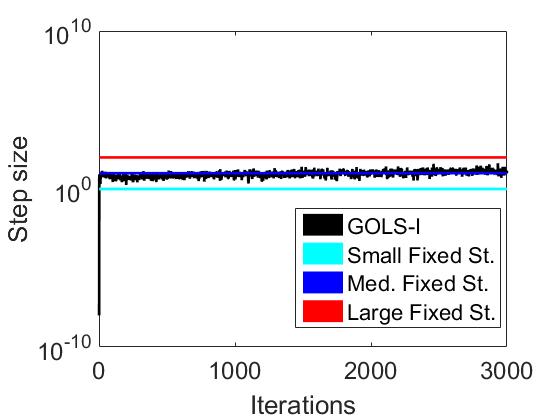}
		\caption{Forest dataset}
	\end{subfigure}%
	
	\begin{subfigure}{.33\textwidth}
		\includegraphics[width=0.95\linewidth]{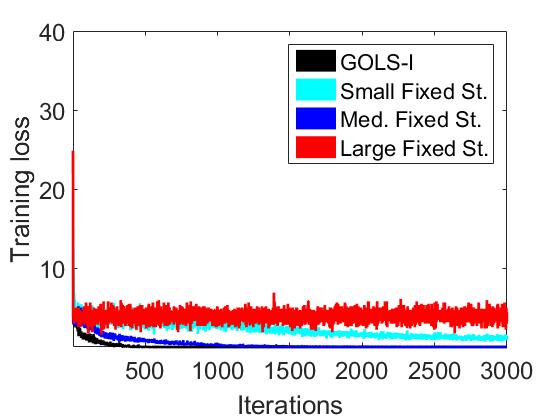}
		\caption{Simulation Failure dataset}
	\end{subfigure}%
	\begin{subfigure}{.33\textwidth}
		\includegraphics[width=0.95\linewidth]{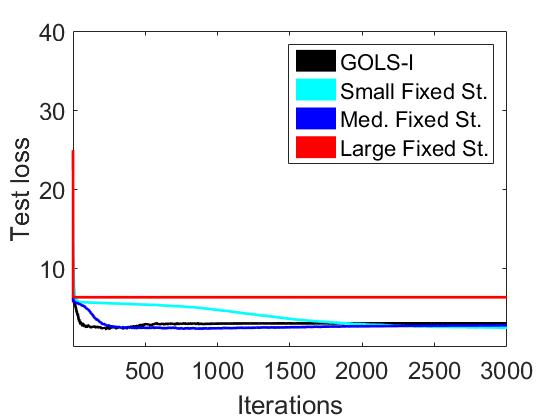}
		\caption{Simulation Failure dataset}
	\end{subfigure}%
	\begin{subfigure}{.33\textwidth}
		\includegraphics[width=0.95\linewidth]{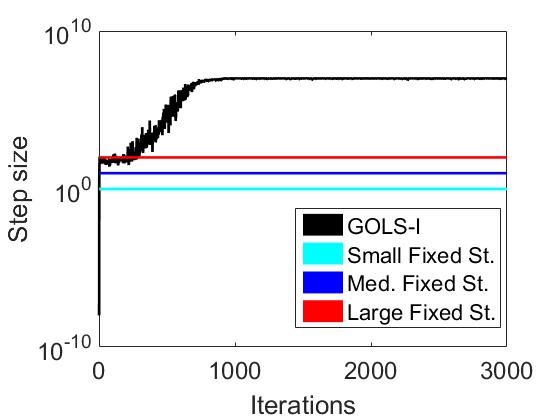}
		\caption{Simulation Failure dataset}
	\end{subfigure}%
	
	\begin{subfigure}{.33\textwidth}
		\includegraphics[width=0.95\linewidth]{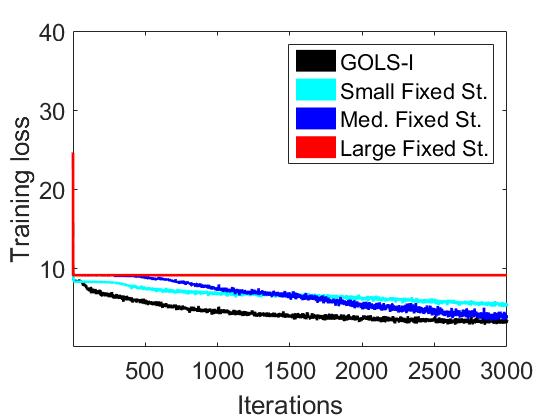}
		\caption{Sensorless Drive dataset}
	\end{subfigure}%
	\begin{subfigure}{.33\textwidth}
		\includegraphics[width=0.95\linewidth]{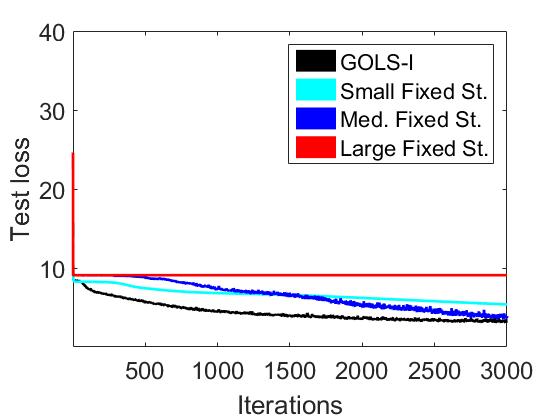}
		\caption{Sensorless Drive dataset}
	\end{subfigure}%
	\begin{subfigure}{.33\textwidth}
		\includegraphics[width=0.95\linewidth]{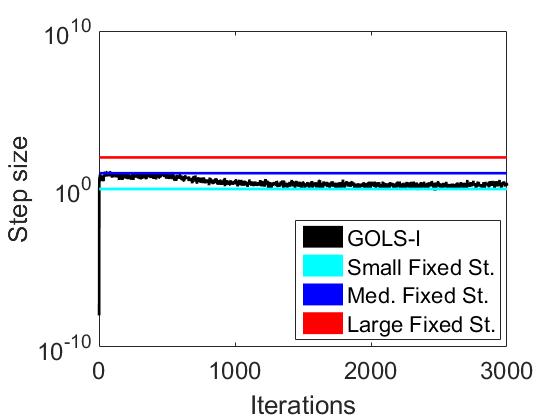}
		\caption{Sensorless Drive dataset}
	\end{subfigure}%
	
	\caption{LS-SGD employed with fixed-step sizes and GOLS-I, as applied to (a)-(c) the "Forests" \citep{Johnson2012}, (d)-(f) "Sensorless Drive" \citep{Paschke2013} and (g)-(i) "Simulation Failure" \citet{Lucas2013} datasets. GOLS-I successfully adapts step sizes according to the characteristics of the problem.}
	\label{fig_adaDatasets}
\end{figure}

Firstly, we consider the training performance of a single algorithm with the selected fixed-step sizes as well as GOLS-I. Hence we choose LS-SGD and train on the "Forests" \citep{Johnson2012,Chevallier2018}, "Sensorless Drive" \citep{Paschke2013,Wenkel2018} and  "Simulation Failure" \citet{Lucas2013,Bischl2017} datasets. We show these particular datasets, as they are representative examples of performances observed over all datasets. The average training and test losses, as well as the step sizes over 10 training runs, are shown in Figure~\ref{fig_adaDatasets}. The fixed-step sizes represent the desired range, from slow to divergent training, while the medium step size is the best performer of the fixed-step sizes. Note that the performance of GOLS-I is competitive or superior to that of the medium step size for the considered problems. However, the step sizes determined by GOLS-I have different characteristics for each example, respectively.

In the case of the Forests dataset in Figure~\ref{fig_adaDatasets}(a)-(c), GOLS-I's determined a step sizes closely approximate the medium fixed step. This was the most commonly observed trend, with 17 out of 22 datasets demonstrating step sizes close to that of the medium step, or starting at the medium step size and slowly tending towards the larger step size as training progresses. Recall, that the initial guess for the step size in the first iteration of GOLS-I is a very conservative $\alpha_{0,0} = 1e-8$. The given results show that GOLS-I adapts the step size from its initial conservative guess to the same magnitude as a manually chosen competitive fixed-step size from the first iteration.

There are other instances, where GOLS-I continues to grow step sizes until the upper limit of allowed step sizes is reached. One out of 3 such cases is the Simulation Failure dataset in Figure~\ref{fig_adaDatasets}(d)-(f). The performance of GOLS-I is marginally superior to that of the medium fixed step, while determining step sizes that are closer to the large fixed step at the beginning of training. However, due to the adaptive nature of GOLS-I (possibly influenced by its ramping up from a small initial guess), it does not diverge like the large fixed step. As training progresses for this problem, and the algorithm approaches an optimum, the gradient magnitude diminishes, which causes GOLS-I to increase the step sizes in search of an SNN-GPP. This is a consequence of decreasing gradient magnitude and the update step formulation proposed in Equation (\ref{eq_updated}).

The performance of GOLS-I for the Sensorless Drive dataset in Figure~\ref{fig_adaDatasets} is particularly interesting. Here, the step sizes determined by GOLS-I start close to the medium fixed step and decrease gradually to the small step size. This occurred for 2 out of the 22 considered problems. In this case, the training performance of GOLS-I is superior to both the medium and the small step sizes. This offers confirmation that GOLS-I is capable of automatically adapting the step size to the requirements of the problem. Conversely, it would be infeasible to pre-empt the shape of the learning rate schedules, such as determined by GOLS-I, for each of the shown problems.

\subsubsection{GOLS-I adapting to different algorithms of a fixed dataset}
\label{sec_golsAlgs}

\begin{figure}[h]
	\begin{subfigure}{.33\textwidth}
		\includegraphics[width=0.95\linewidth]{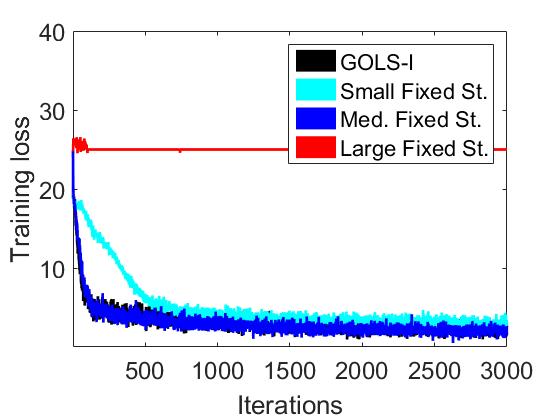}
		\caption{LS-NAG}
	\end{subfigure}%
	\begin{subfigure}{.33\textwidth}
		\includegraphics[width=0.95\linewidth]{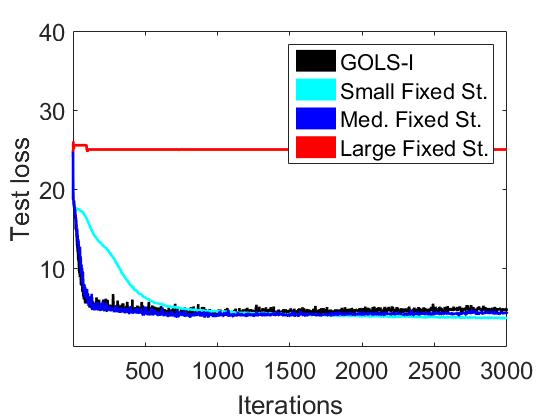}
		\caption{LS-NAG}
	\end{subfigure}%
	\begin{subfigure}{.33\textwidth}
		\includegraphics[width=0.95\linewidth]{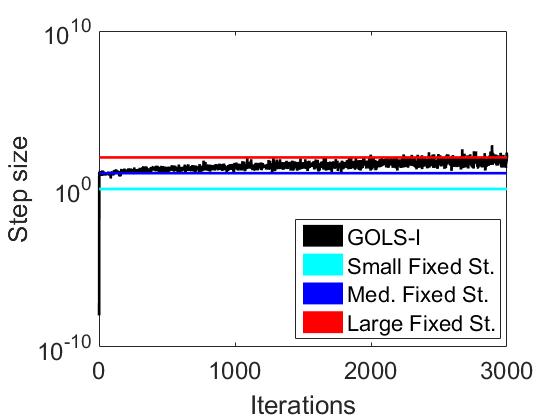}
		\caption{LS-NAG}
	\end{subfigure}%
	
	\begin{subfigure}{.33\textwidth}
		\includegraphics[width=0.95\linewidth]{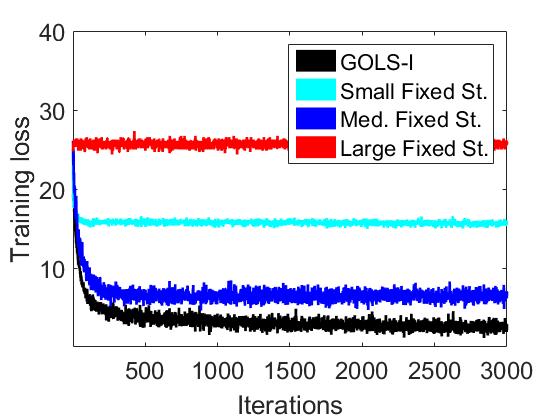}
		\caption{LS-Adadelta}
	\end{subfigure}%
	\begin{subfigure}{.33\textwidth}
		\includegraphics[width=0.95\linewidth]{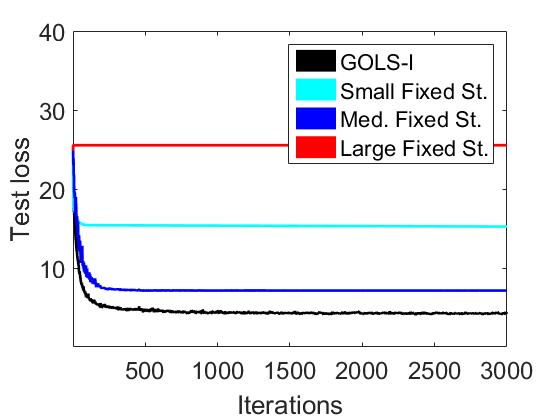}
		\caption{LS-Adadelta}
\end{subfigure}%
	\begin{subfigure}{.33\textwidth}
		\includegraphics[width=0.95\linewidth]{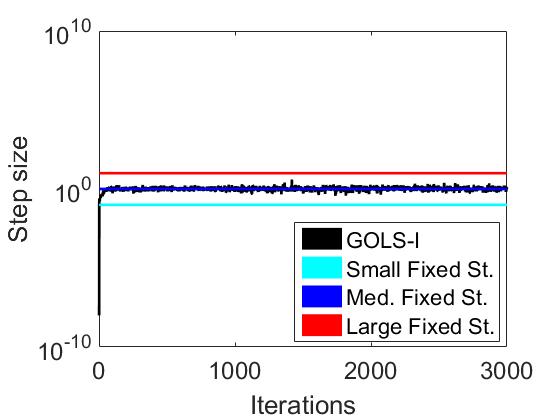}
		\caption{LS-Adadelta}
	\end{subfigure}%
	
	\begin{subfigure}{.33\textwidth}
		\includegraphics[width=0.95\linewidth]{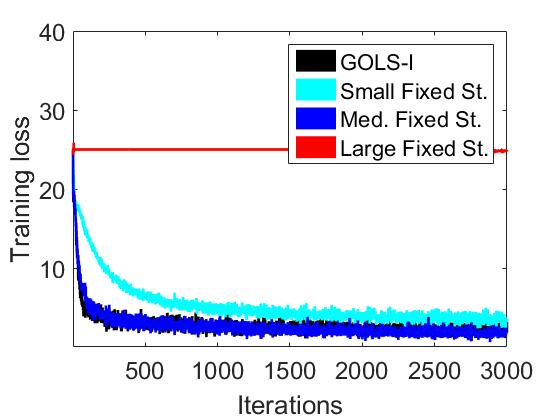}
		\caption{LS-Adam}
	\end{subfigure}%
	\begin{subfigure}{.33\textwidth}
		\includegraphics[width=0.95\linewidth]{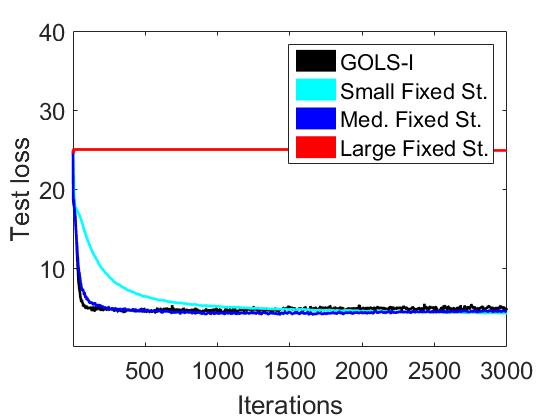}
		\caption{LS-Adam}
	\end{subfigure}%
	\begin{subfigure}{.33\textwidth}
		\includegraphics[width=0.95\linewidth]{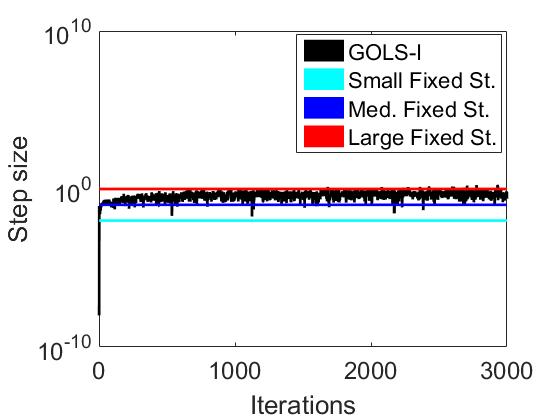}
		\caption{LS-Adam}
	\end{subfigure}%
	\caption{The "Forests" \citep{Johnson2012} problem as trained using (a)-(c) LS-NAG, (d)-(f) LS-Adadelta and (g)-(i) LS-Adam with both GOLS-I and fixed-step sizes. GOLS-I automatically recovers step sizes close to medium fixed steps for the shown algorithms, while each algorithm has different medium fixed-step sizes, each an order of magnitude apart.}
	\label{fig_adaAlgs}
\end{figure}

Subsequently, we restrict the investigation to the "Forests" \citep{Johnson2012} dataset, while considering the LS-NAG, LS-Adadelta and LS-Adam algorithms with their respective fixed-step sizes and GOLS-I implementations. The training and test dataset losses with corresponding step sizes are given in Figure~\ref{fig_adaAlgs}. GOLS-I NAG performs similarly to the case considered for the same dataset as applied to GOLS-I SGD, see Figure~\ref{fig_adaDatasets}(a)-(c). This is the most common case, where GOLS-I determines step sizes close to the medium fixed step during the early stages of training, while tending towards the large fixed step towards the latter half. Here, the training performance of GOLS-I is indistinguishable from that of the medium fixed step.

The interaction between GOLS-I and LS-Adadelta is noteworthy. Adadelta does not contain a learning rate parameter. However, we introduced the scalar modifier, $\alpha_{n,I_n} = 1$, to its formulation to explore the characteristics of its search directions. Unsurprisingly, LS-Adadelta's best performance with fixed-step sizes is when $\alpha_{n,I_n} = 1$, i.e. recovering its original formulation. However, the improvement in performance of GOLS-I Adadelta is significant: The GOLS-I recovers average step sizes close to 1, while its performance is superior to that of fixed-step size 1 in both training and test losses. This implies that the small deviations from the medium step sizes according to updates performed with GOLS-I improve the adaptation of LS-Adadelta to the characteristics of the encountered loss surface.

GOLS-I Adam returns similar performance characteristics to those of GOLS-I NAG, with the step size increasing from the medium step size as training progresses. However, the medium fixed-step sizes for LS-NAG, LS-Adadelta and LS-Adam are all an order of magnitude apart at 10, 1 and 0.1, respectively. This means that GOLS-I can automatically adapt the determined step size according to the characteristics of the training algorithm dictating the search direction, while resulting in superior or competitive training performance.

\subsubsection{Averaged comparison of algorithms over 22 datasets}

\begin{figure}[h]
	\begin{subfigure}{.33\textwidth}
		\includegraphics[width=0.95\linewidth]{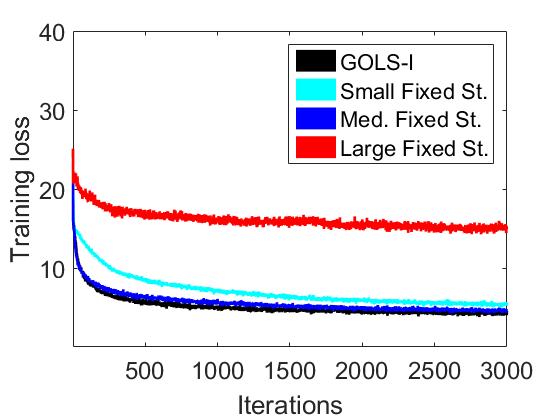}
		\caption{LS-SGD}
	\end{subfigure}%
	\begin{subfigure}{.33\textwidth}
		\includegraphics[width=0.95\linewidth]{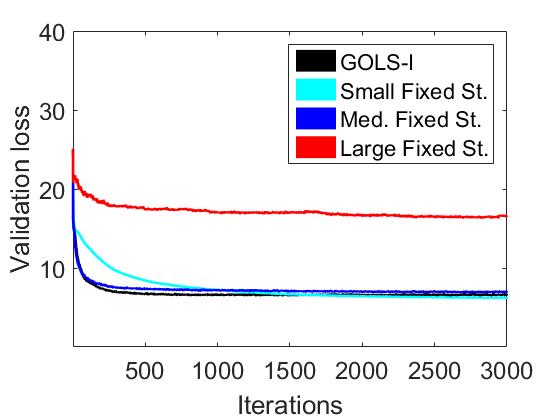}
		\caption{LS-SGD}
	\end{subfigure}%
	\begin{subfigure}{.33\textwidth}
		\includegraphics[width=0.95\linewidth]{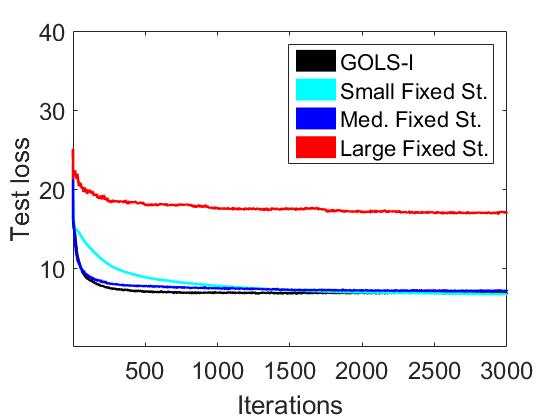}
		\caption{LS-SGD}
	\end{subfigure}%

	\begin{subfigure}{.33\textwidth}
		\includegraphics[width=0.95\linewidth]{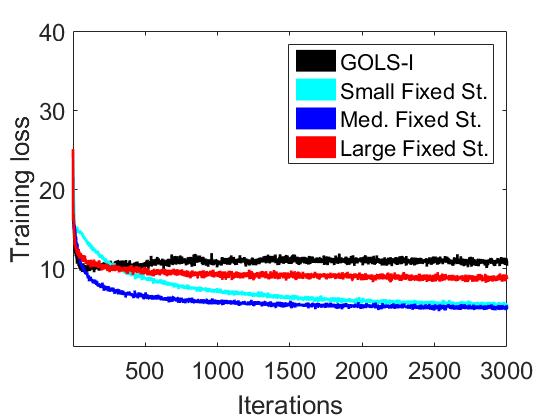}
		\caption{LS-SGDM}
	\end{subfigure}%
	\begin{subfigure}{.33\textwidth}
		\includegraphics[width=0.95\linewidth]{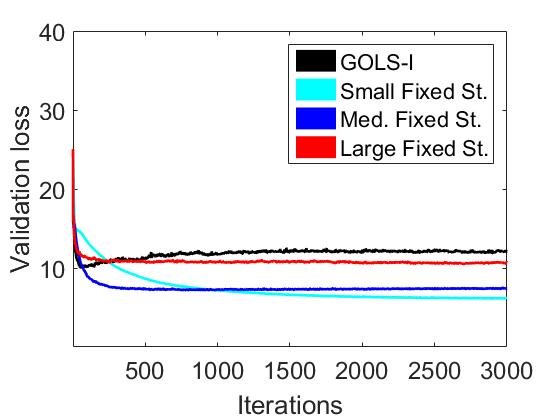}
		\caption{LS-SGDM}
	\end{subfigure}%
	\begin{subfigure}{.33\textwidth}
		\includegraphics[width=0.95\linewidth]{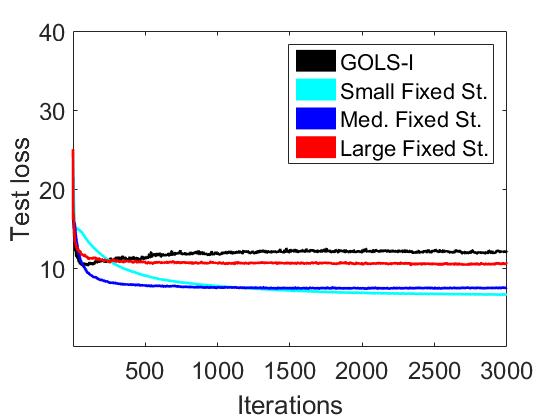}
		\caption{LS-SGDM}
	\end{subfigure}%

	\begin{subfigure}{.33\textwidth}
		\includegraphics[width=0.95\linewidth]{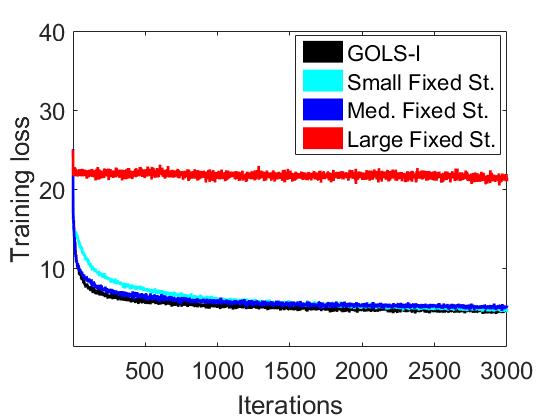}
		\caption{LS-NAG}
	\end{subfigure}%
	\begin{subfigure}{.33\textwidth}
		\includegraphics[width=0.95\linewidth]{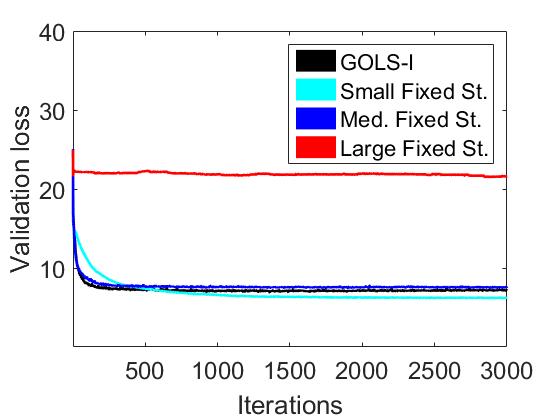}
		\caption{LS-NAG}
	\end{subfigure}%
	\begin{subfigure}{.33\textwidth}
		\includegraphics[width=0.95\linewidth]{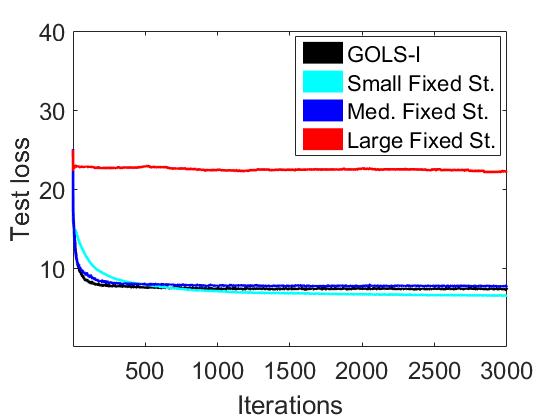}
		\caption{LS-NAG}
	\end{subfigure}%
	
	\caption{The training, test and validation dataset errors for (a)-(c) LS-SGD, (d)-(f) LS-SGD with momentum and (g)-(i) LS-NAG, averaged over all datasets for 10 different starting points per analysis. The algorithms are implemented with different fixed-step sizes and the gradient-only based GOLS-I.}
	\label{fig_err_ave_indivs1}
\end{figure}

\begin{figure}[h]	
	\begin{subfigure}{.33\textwidth}
		\includegraphics[width=0.95\linewidth]{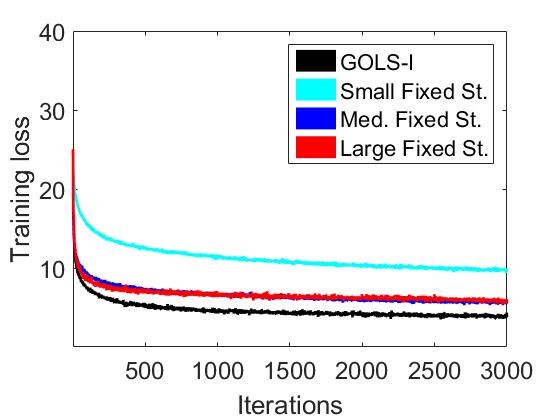}
		\caption{LS-Adagrad}
	\end{subfigure}%
	\begin{subfigure}{.33\textwidth}
		\includegraphics[width=0.95\linewidth]{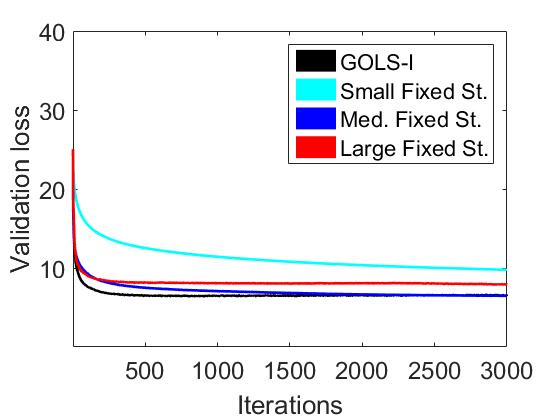}
		\caption{LS-Adagrad}
	\end{subfigure}%
	\begin{subfigure}{.33\textwidth}
		\includegraphics[width=0.95\linewidth]{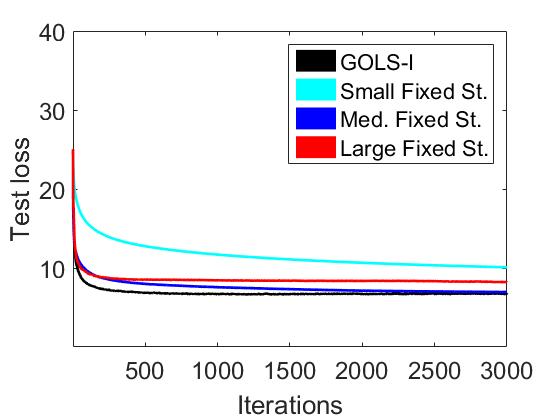}
		\caption{LS-Adagrad}
	\end{subfigure}%

	\begin{subfigure}{.33\textwidth}
		\includegraphics[width=0.95\linewidth]{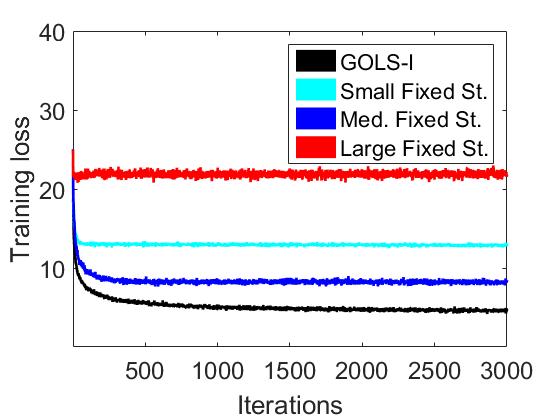}
		\caption{LS-Adadelta}
	\end{subfigure}%
	\begin{subfigure}{.33\textwidth}
		\includegraphics[width=0.95\linewidth]{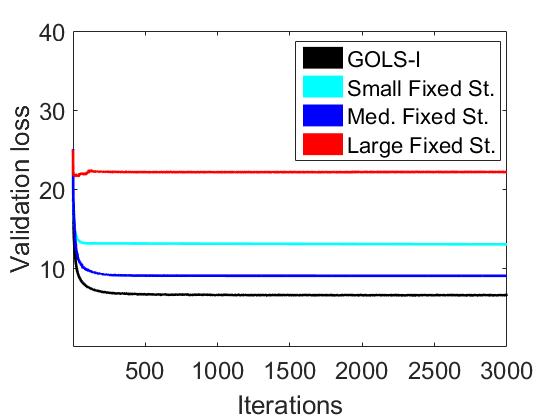}
		\caption{LS-Adadelta}
	\end{subfigure}%
	\begin{subfigure}{.33\textwidth}
		\includegraphics[width=0.95\linewidth]{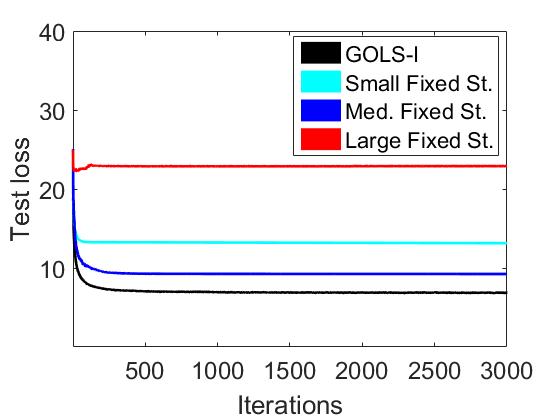}
		\caption{LS-Adadelta}
	\end{subfigure}%

	\begin{subfigure}{.33\textwidth}
		\includegraphics[width=0.95\linewidth]{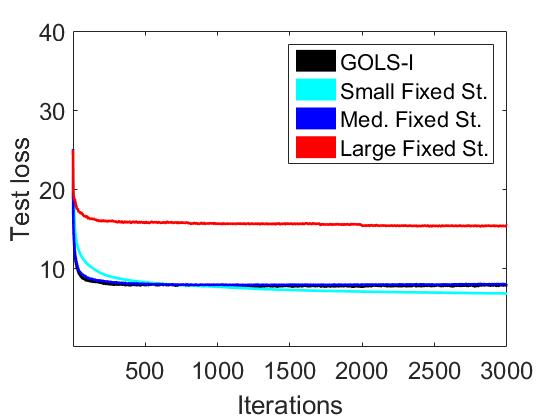}
		\caption{LS-Adam}
	\end{subfigure}%
	\begin{subfigure}{.33\textwidth}
		\includegraphics[width=0.95\linewidth]{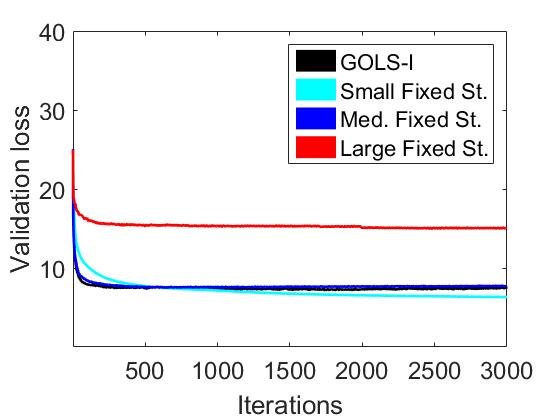}
		\caption{LS-Adam}
	\end{subfigure}%
	\begin{subfigure}{.33\textwidth}
		\includegraphics[width=0.95\linewidth]{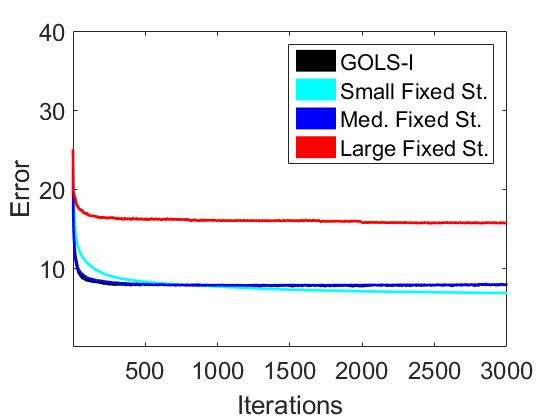}
		\caption{LS-Adam}
	\end{subfigure}%
	
	\caption{The training, test and validation dataset errors for (a)-(c) LS-Adagrad, (d)-(f) LS-Adadelta and (g)-(i) LS-Adam, averaged over all datasets for 10 different starting points per analysis. The algorithms are implemented with different fixed-step sizes and the gradient-only based GOLS-I.}
	\label{fig_err_ave_indivs2}
\end{figure}

Now that it has been established that GOLS-I fulfils its intended purpose for independent datasets and algorithms, we consider the average performance of all 6 algorithms over the 22 considered problem to examine the generality of the method. The resulting averaged training, validation and test dataset losses are shown in Figures~\ref{fig_err_ave_indivs1} and \ref{fig_err_ave_indivs2}. As demonstrated, the fixed-step sizes encompass the performance range from too slow, to divergent within 3 orders of magnitude. For our implementation of LS-Adagrad, the largest fixed step does not diverge, but instead results in the worst generalization of the 3 step sizes.

For most algorithms with the use of GOLS-I, the average validation and test set losses show that overfitting occurs after approximately 1000 iterations for the datasets considered in this study. Importantly, the characteristics of the validation and test loss plots are comparable. This means that the network designer can reliably use the behaviour of the validation dataset to enforce a stopping criterion or tune other network parameters.

As demonstrated in Sections~\ref{sec_golsData} and \ref{sec_golsAlgs}, the training loss for GOLS-I is mostly either the best performer or competitive with the medium fixed step. The obvious outlier is LS-SGDM, where GOLS-I performs worse than any fixed step. This is a result of LS-SGDM adding momentum, which is a fixed fraction of the previous update after the line search has already determined the step size. This forces the optimizer past minima and up ascent directions. For this reason, LS-SGDM is ill-suited to be used with a line search such as GOLS-I. 

Conversely, small fixed-step sizes result in update steps with incremental changes in the loss function, which also translates to less aggressive input from the momentum term, allowing the algorithm to more closely follow the contours of the loss function. The ascending behaviour of GOLS-I SGDM can be addressed by reducing the parameter $\gamma_m$, shifting the algorithm's behaviour closer to LS-SGD. The adaptation to the momentum algorithm offered by LS-NAG is useful in overcoming the ascending problem, resulting in performance comparable to LS-SGD for the coupled directions class of algorithms.

In terms of the uncoupled class of algorithms, GOLS-I performed best with LS-Adagrad and LS-Adadelta relative to fixed-step sizes. GOLS-I Adagrad produced an average performance that was superior to that of the medium fixed step, while GOLS-I Adadelta on average performed similarly to the results explored in Section~\ref{sec_golsAlgs}. As discussed, this is an indication that the bottleneck for LS-Adadelta is not the search direction, but rather the step size along its search direction. LS-Adam, in this case, is the outlier, with the smallest fixed step performing best as training extends beyond approx 1000 iterations. This is due to LS-Adam's formulation, which includes a term that acts like momentum. However, this effect is investigated more clearly in Section~\ref{sec_deepHL}, considering deep networks.

\subsubsection{Average performance over all algorithms}

\begin{figure}[h!]
	\centering
	\begin{subfigure}{.33\textwidth}
		\centering 
		\includegraphics[width=0.95\linewidth]{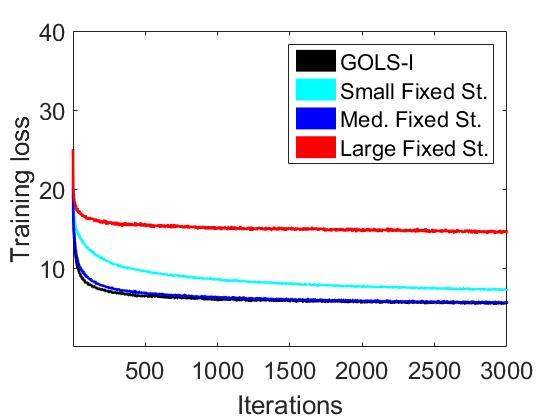}
		\caption{}
	\end{subfigure}%
	\begin{subfigure}{.33\textwidth}
		\centering
		\includegraphics[width=0.95\linewidth]{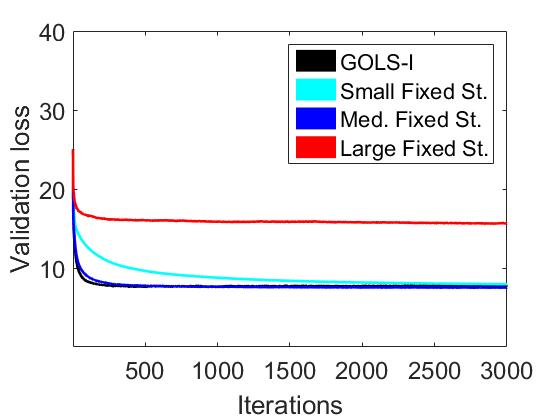}
		\caption{}
	\end{subfigure}%
	\begin{subfigure}{.33\textwidth}
		\centering 
		\includegraphics[width=0.95\linewidth]{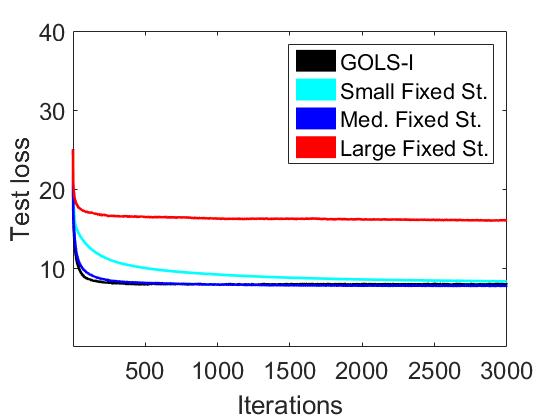}
		\caption{}
	\end{subfigure}%
	
	\begin{subfigure}{.33\textwidth}
		\centering 
		\includegraphics[width=0.95\linewidth]{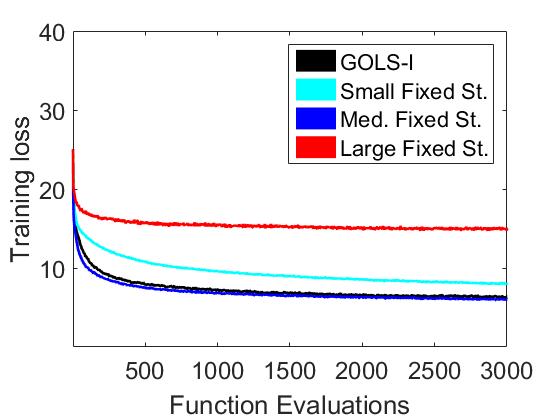}
		\caption{}
	\end{subfigure}%
	\begin{subfigure}{.33\textwidth}
		\centering
		\includegraphics[width=0.95\linewidth]{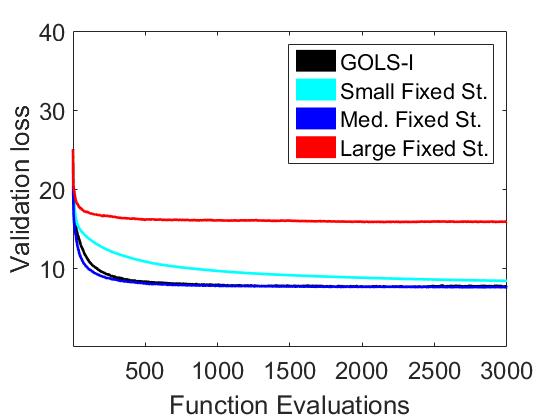}
		\caption{}
	\end{subfigure}%
	\begin{subfigure}{.33\textwidth}
		\centering 
		\includegraphics[width=0.95\linewidth]{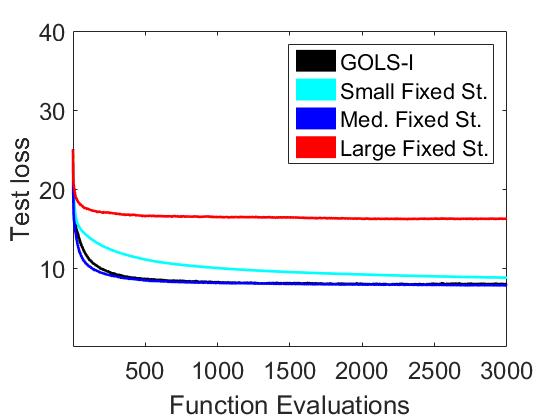}
		\caption{}
	\end{subfigure}%
	\caption{Average training, test and validation dataset errors over all datasets and all algorithm in terms of (a)-(c) iterations and (d)-(f) function evaluations. For GOLS-I, the number of function evaluations per iteration is on average 2.5, whereas that of a fixed-step size approach is always 1. This is also a proxy for computational cost. Therefore, plotting in terms of function evaluations corrects for this disparity. Though GOLS-I is a factor of 2.5 more expensive than the medium fixed step on average, it remains competitive, while requiring no tuning.}
	\label{fig_err_ave_all-itsVfv}
\end{figure}

We consider the average performance over all datasets and all algorithms in Figure~\ref{fig_err_ave_all-itsVfv}. We do this to highlight two aspects: 1) How GOLS-I generalizes over all considered dataset-algorithm combinations and 2) How GOLS-I competes overall in terms of function evaluations per iteration compared to fixed-step sizes.

The results up to and including Figures~\ref{fig_err_ave_all-itsVfv}(a)-(c) are given in terms of iterations. This is done to compare per iteration performance gains and step sizes between GOLS-I and the fixed steps. On a per iteration comparison, GOLS-I is competitive with or superior to the selected medium step size.

As a line search, GOLS-I performs several function evaluations per iteration. For this study, a function evaluation is considered to be an evaluation of $\tilde{\boldsymbol{g}}(\boldsymbol{x})$. A constant step size uses only a single function evaluation per iteration. On average GOLS-I performed between 1-4 function evaluations per iteration. Thus to compare the relative computational cost of GOLS-I compared to fixed-step sizes, the average algorithm performances of the constant step methods and GOLS-I are plotted in terms of function evaluations in Figures~\ref{fig_err_ave_all-itsVfv}(d)-(f). Although GOLS-I is on average a factor of 2.5 more expensive compared to a well-chosen medium step size, the extra cost does not offset the benefit gained by not requiring to search for adequate fixed-step sizes.


\subsection{Multiple hidden layer neural networks}

\label{sec_deepHL}

Next, we extend our analyses to include deep neural networks. We consider GOLS-I with LS-SGD, LS-Adagrad and LS-Adam; each applied to neural networks with an increasing numbers of hidden layers \citep{Bengio2009}, ranging from 1 to 6. All parameters remained consistent with those of the study described in Section \ref{sec_SHLNN}, except for the number of hidden layers. This investigation omitted the larger datasets, namely: Simulation failures \citep{Lucas2013}, Defaulted credit cards \citep{Yeh2009}, Sensorless drive diagnosis \citep{Paschke2013} and MNIST \citep{Lecun1998}, which makes a total number of datasets 18 for this section.

\begin{figure}[h!]
	\begin{subfigure}{.33\textwidth}
		\includegraphics[width=0.95\linewidth]{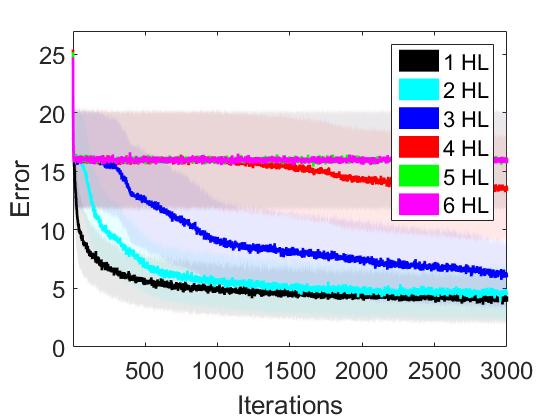}
		\caption{GOLS-I SGD}
	\end{subfigure}%
	\begin{subfigure}{.33\textwidth}
		\includegraphics[width=0.95\linewidth]{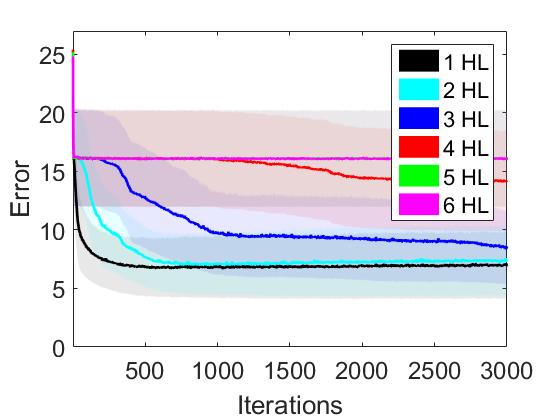}
		\caption{GOLS-I SGD}
	\end{subfigure}%
	\begin{subfigure}{.33\textwidth}
		\includegraphics[width=0.95\linewidth]{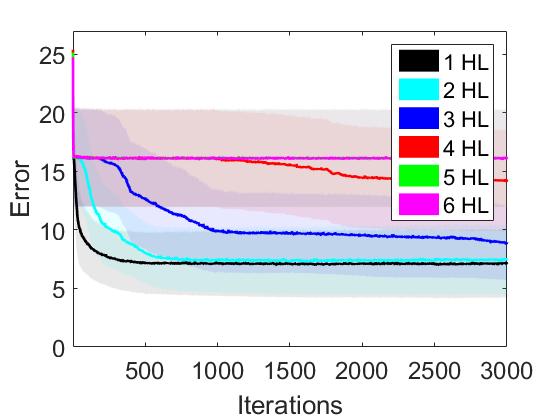}
		\caption{GOLS-I SGD}
	\end{subfigure}%

	\begin{subfigure}{.33\textwidth}
		\includegraphics[width=0.95\linewidth]{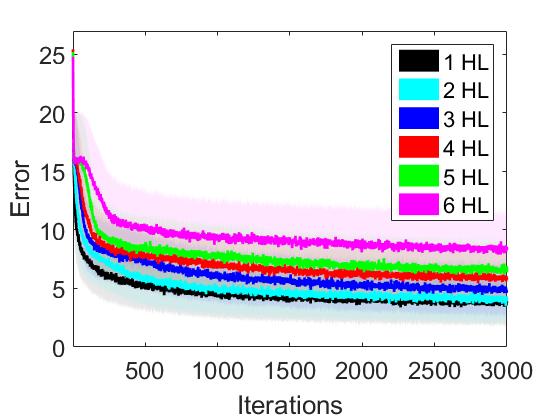}
		\caption{GOLS-I Adagrad}
	\end{subfigure}%
	\begin{subfigure}{.33\textwidth}
		\includegraphics[width=0.95\linewidth]{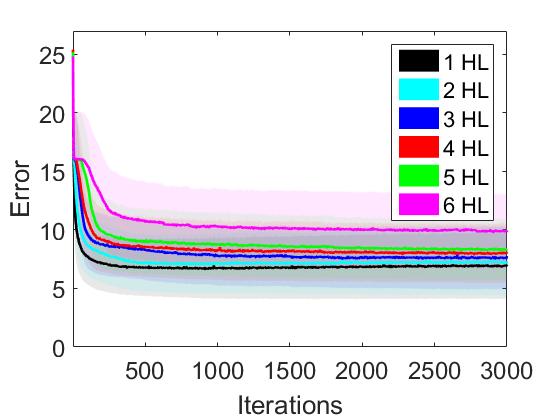}
		\caption{GOLS-I Adagrad}
	\end{subfigure}%
	\begin{subfigure}{.33\textwidth}
		\includegraphics[width=0.95\linewidth]{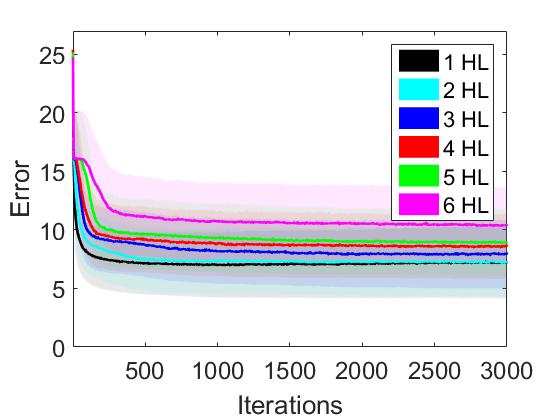}
		\caption{GOLS-I Adagrad}
	\end{subfigure}%

	\begin{subfigure}{.33\textwidth}
		\includegraphics[width=0.95\linewidth]{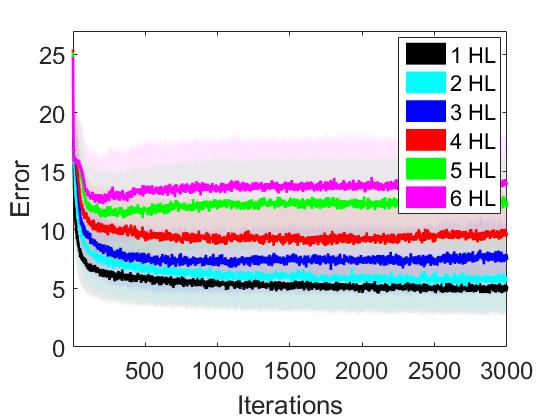}
		\caption{GOLS-I Adam}
	\end{subfigure}%
	\begin{subfigure}{.33\textwidth}
		\includegraphics[width=0.95\linewidth]{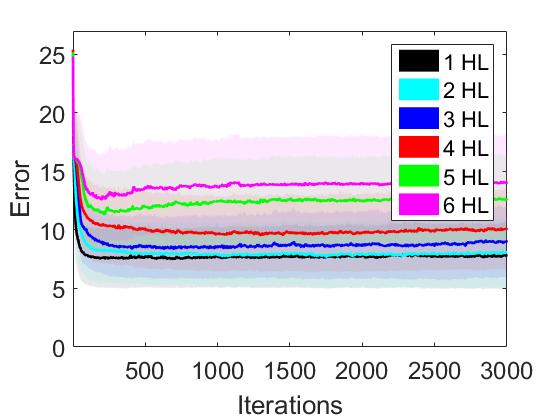}
		\caption{GOLS-I Adam}
	\end{subfigure}%
	\begin{subfigure}{.33\textwidth}
		\includegraphics[width=0.95\linewidth]{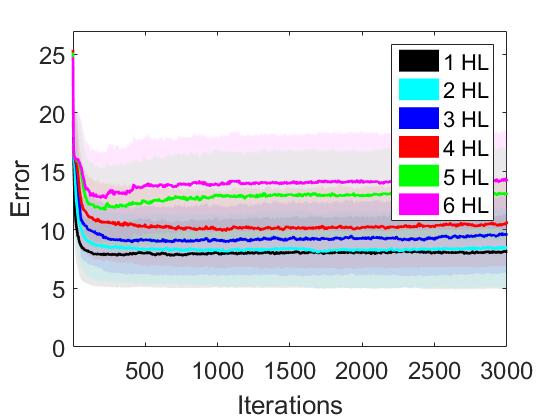}
		\caption{GOLS-I Adam}
	\end{subfigure}%

	\begin{subfigure}{.33\textwidth}
		\includegraphics[width=0.95\linewidth]{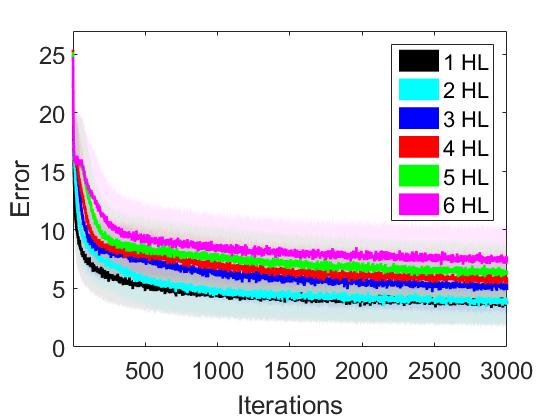}
		\caption{GOLS-I Adam, $\beta_1=0$}
	\end{subfigure}%
	\begin{subfigure}{.33\textwidth}
		\includegraphics[width=0.95\linewidth]{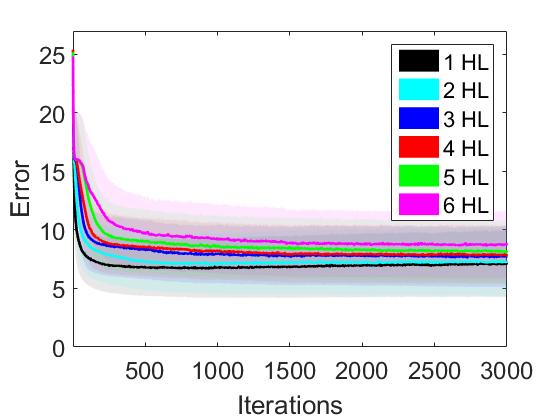}
		\caption{GOLS-I Adam, $\beta_1=0$}
	\end{subfigure}%
	\begin{subfigure}{.33\textwidth}
		\includegraphics[width=0.95\linewidth]{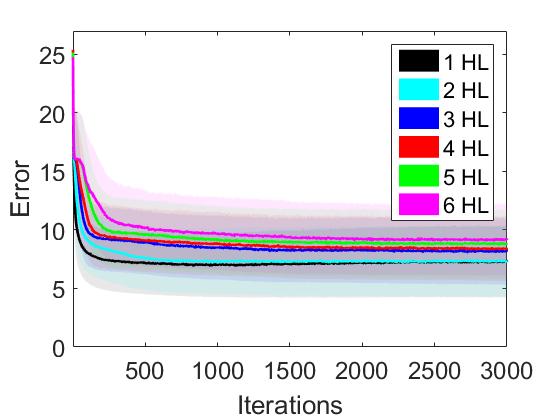}
		\caption{GOLS-I Adam, $\beta_1=0$}
	\end{subfigure}%

	\caption{Training, test and validation dataset losses of (a)-(c) GOLS-I SGD, (d)-(f) GOLS-I Adagrad, (g)-(i) GOLS-I Adam and (j)-(l) GOLS-I Adam with $\beta_1=0$ for 1-6 of hidden layers. The averages are taken over a subset of datasets with 10 separate training runs per dataset.}
	\label{fig_deepHL_all}
\end{figure}

Figures~\ref{fig_deepHL_all}(a)-(c) show the average training, validation and test losses over 1 to 6 hidden layers using GOLS-I SGD. Here, variance clouds are shown with shaded colours around the mean, which is denoted by a solid line. Reasonable training performance can be achieved for smaller networks of 1 to 3 hidden layers in 3000 iterations. Convergence is slow on networks with 4 hidden layers, although some progress is made within 3000 iterations. However, there is a severe decrease in performance for networks with 5 and 6 hidden layers, with higher variance between training runs, compared to smaller networks. This indicates, that the diminishing gradient problem \citep{Hochreiter2009} begins to affect the optimization algorithm considerably from this network depth onwards. This plot is representative not only for GOLS-I SGD, but also for analyses not shown here concerning GOLS-I SGDM and the GOLS-I NAG algorithms, indicating the difficulty present for the coupled directions class of algorithms for this problem. It is also to be noted, that this problem can be alleviated with more advanced network initialization methods, as proposed by \cite{Glorot2010}.

The corresponding results with the use of GOLS-I Adagrad are shown in Figures~\ref{fig_deepHL_all}(d)-(f). As a representative of the uncoupled directions class of algorithms, GOLS-I Adagrad performs much more consistently across increasing numbers of hidden layers, exhibiting lower variance between training runs. However, there is still a slight relative decrease in performance for deeper networks. The test and validation error plots in Figures~\ref{fig_deepHL_all}(e) and (f) show that validation and test dataset losses increase for the smaller networks due to overfitting, while the larger networks still progress towards a minimum, albeit at a slower convergence rate. 
 
In the case of GOLS-I Adam, as shown in Figures~\ref{fig_deepHL_all}(g)-(i), convergence occurs at a slower rate than with GOLS-I Adagrad for deeper networks. It is here, that the momentum-like term in LS-Adam hampers its performance with GOLS-I considerably. This term, $\hat{\boldsymbol{m}}_{n+1}$ in the numerator of its formulation, see Appendix \ref{sec_LS-Adam}, is governed by the $\beta_1$ parameter. By default, this parameter is set as $\beta_1 = 0.9$. However, if this parameter is reduced to $\beta_1 = 0$, the momentum behaviour of LS-Adam is deactivated, while the denominator still contributes to producing uncoupled, scaled directions. In Figures~\ref{fig_deepHL_all}(j)-(l), the results of separate training runs of GOLS-I Adam with $\beta_1 = 0$ are shown. It is clear, that there is a vast improvement in performance, with performance now being comparable to that of GOLS-I Adagrad.

This study has confirmed that generally, momentum in the context of GOLS-I should be avoided. And secondly, that the element-wise scaling of the direction entries creates a total direction which allows for greater perturbation in the deeper layers of the networks, promoting faster training. The inclusion of GOLS-I into the considered training algorithm capitalizes on their various search directions for efficient, deep network training, while requiring no parameter tuning.

\subsection{Cifar10 with ResNet18}


Now that GOLS-I has been demonstrated on a variety of foundational problems, we apply it to an architecture more likely to be encountered in practice. We therefore consider the Cifar10 classification dataset with the ResNet18 Convolutional Neural Network (CNN) architecture \citep{He2016}. The baseline PyTorch code for this problem was sourced from \citet{Kuangliu2018}, which was then modified to accommodate GOLS-I. This experiment makes use of the cross-entropy loss for $\ell(\boldsymbol{x},\boldsymbol{t}_b)$, which is more popular for classification tasks, as it can be viewed as minimizing the KL divergence between the distributions of the neural network, and the target outputs \citep{GoodfellowBengioCourville2016}. The algorithms considered with GOLS-I are LS-SGD, LBFGS \cite{Nocedal1980} (as used with a line search as  "LS-LBFGS"), LS-Adagrad and LS-Adam as adapted for PyTorch 1.0. Incorporating lessons learnt from Section \ref{sec_deepHL}, the momentum term in LS-Adam is turned off, i.e. $\beta_1 = 0$. The batch size chosen for this problem was constant at 128, and training was limited to $40,000$ iterations. In this case, $\mathcal{B}_{n,i}$ is randomly sampled from the pool of training data without replacement, until all training observations have been sampled. Thereafter, the training dataset is shuffled, and the process repeats.

\begin{figure}[h!]
	\centering
	\begin{subfigure}{.49\textwidth}
		\centering 
		\includegraphics[width=0.85\linewidth]{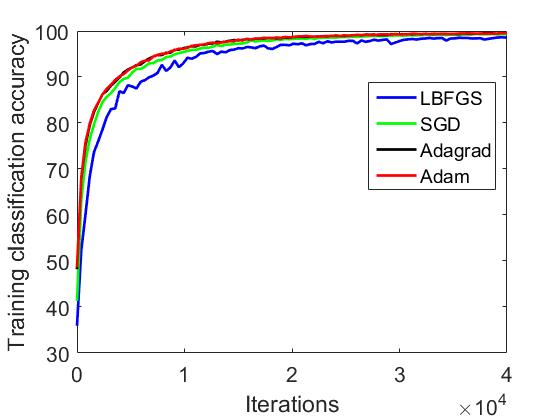}
		\caption{Training classification accuracy}
	\end{subfigure}%
	\begin{subfigure}{.49\textwidth}
		\centering
		\includegraphics[width=0.85\linewidth]{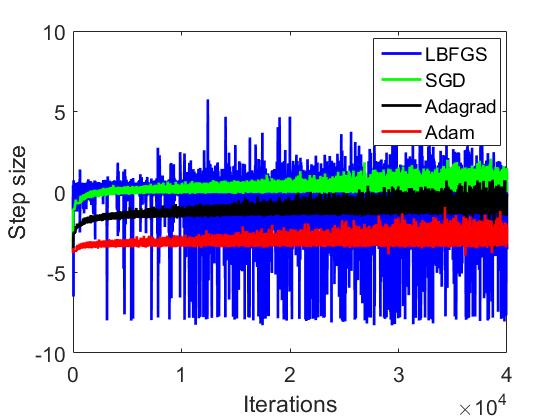}
		\caption{Step sizes}
	\end{subfigure}%
	
	\begin{subfigure}{.49\textwidth}
		\centering 
		\includegraphics[width=0.85\linewidth]{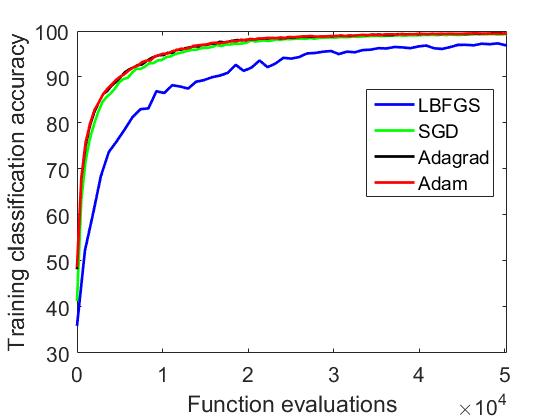}
		\caption{Performance versus cost}
	\end{subfigure}%
	\begin{subfigure}{.49\textwidth}
		\centering
		\includegraphics[width=0.85\linewidth]{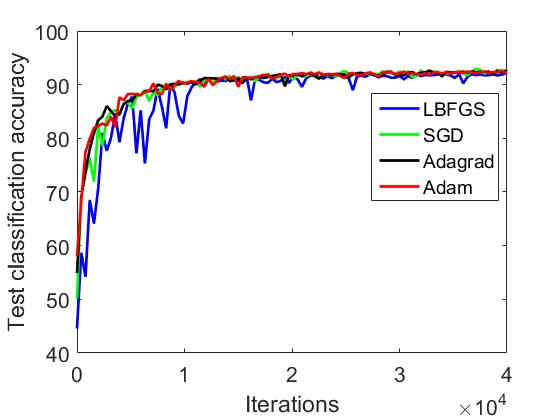}
		\caption{Test classification accuracy}
	\end{subfigure}%
	
	\caption{The (a) training classification accuracy, (b) step sizes, (c) training classification accuracy in terms of function evaluations and (d) test classification accuracy for various training algorithms as applied to the Cifar10 dataset with the ResNet18 architecture. The step sizes are determined by GOLS-I, while the explicit reference to "GOLS-I" is omitted in the legends for compactness. The step sizes are different between algorithms, while their performances per iteration are comparable. This indicates that step sizes and not search directions may be the bottleneck for training this problem. However, the cost of each algorithm is not the same, revealing GOLS-I LBFGS to be the worst performer in terms of computational efficiency.}
	\label{fig_cifar}
\end{figure}

\begin{figure}[h!]
	\centering
	\begin{subfigure}{.33\textwidth}
		\centering 
		\includegraphics[width=0.99\linewidth]{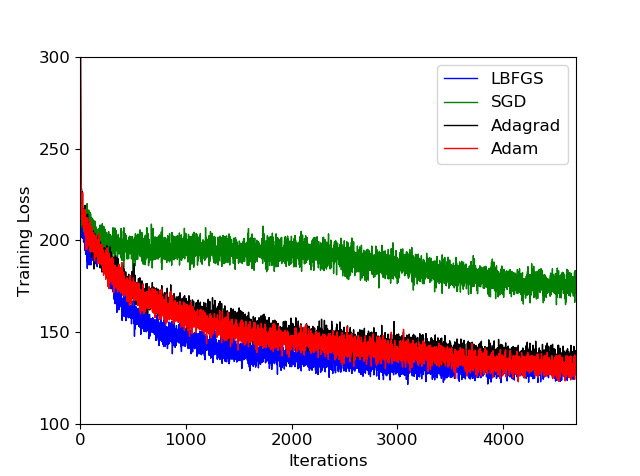}
		\caption{Training loss}
	\end{subfigure}%
	\begin{subfigure}{.33\textwidth}
		\centering
		\includegraphics[width=0.99\linewidth]{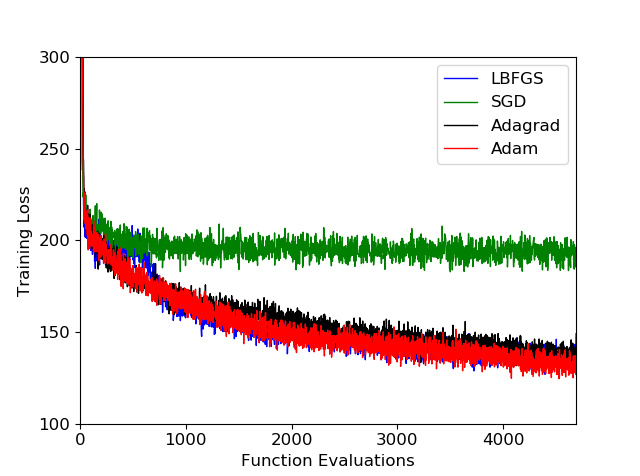}
		\caption{Training loss}
	\end{subfigure}%
	\begin{subfigure}{.33\textwidth}
		\centering
		\includegraphics[width=0.99\linewidth]{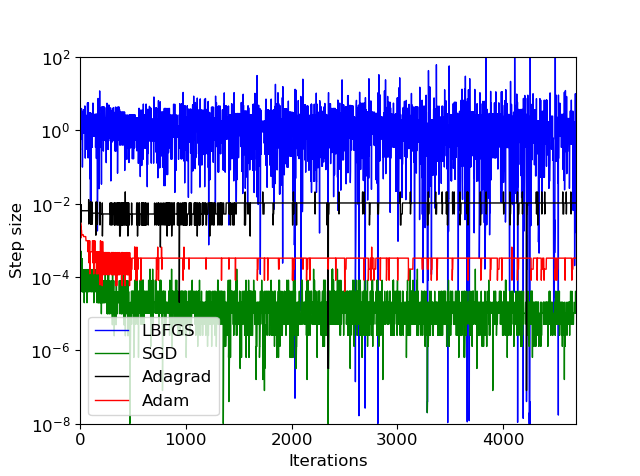}
		\caption{Step sizes}
	\end{subfigure}%
	
	\caption{The (a) training loss in terms of iteration, (b) training loss in terms of function evaluations, (c) step sizes for various training algorithms as applied to variational autoencoder \citep{Zuo2018} training. The step sizes as determined by GOLS-I are again different for every algorithm (with "GOLS-I" omitted in the legends for compactness), while highlighting the effectiveness of Hessian approximations generated by LS-LBFGS.}
	\label{fig_vae}
\end{figure}

The resulting training and test classification accuracy, step sizes and training classification in terms of function evaluations are shown in Figure~\ref{fig_cifar}. The performances of GOLS-I SGD, GOLS-I Adagrad and GOLS-I Adam are almost indistinguishable in training, while marginally outperforming GOLS-I LBFGS. This shows that for this problem, the quality of search directions constructed by GOLS-I LBFGS are inferior to those of the remaining algorithms. The GOLS-I LBFGS directions are generated based Hessian approximations, computed from the gradient information at previous iterations. However, the combination of making use of dynamic MBSS as well as GOLS-I being an inexact line search reduces the quality of the Hessian approximation and can result in the generation of ascent directions after several iterations. In such cases, GOLS-I compensates by taking the minimum step size in that direction, resulting in the high variance in step sizes we observe in Figure~\ref{fig_cifar}(b). This is also confirmed by the relatively noisy training classification errors of GOLS-I LBFGS compared to the remaining algorithms. 

We capture the number of function evaluations performed for each iteration and determine the average number of function evaluations per iteration during training, as summarized in Table~\ref{tbl_cifar_fvals}. 
This shows that GOLS-I LBFGS is on average $77\%$ more expensive in the number of function evaluations performed per iteration than the remaining algorithms. To compare performance versus computational cost between the algorithms, we plot the training classification accuracy in terms of function evaluations in Figure~\ref{fig_cifar}(c). This plot highlights, that the search directions constructed by GOLS-I LBFGS do not offset the cost of their construction for this problem. However, given that this is to our knowledge the first GOLS-I LBFGS formulation that incorporates dynamic MBSS line searches in the construction of its Hessian approximations without any further modifications, the offered performance is not to be discounted.  Future work might consider whether modifications such as increasing batch size or altering parameters in LS-LBFGS can improve its efficiency. However, the insignificant difference in performance between LS-Adagrad, LS-Adam and LS-SGD indicates that the training algorithms are relatively insensitive to the inclusion of curvature information for this problem. Therefore, CIFAR10 with ResNet18 may not be a suitable platform to thoroughly test the validity of LS-LBFGS' Hessian approximations in neural network training.

\begin{table}[h!]
	\centering
	\scalebox{0.75}{
		\begin{tabular}{|c|c|c|c|c|}
			\hline 
			& \textbf{LS-LBFGS} & \textbf{LS-SGD} & \textbf{LS-Adagrad} & \textbf{LS-Adam, $\beta_1 = 0$} \\ 
			\hline 
			\textbf{Average \# function evaluations per update} & $2.35$ & $1.26$ & $1.28$ & $1.28$ \\ 
			\hline 
		\end{tabular}
	} 
	\caption{Algorithmic comparison in terms of cost. A constant number of iteration was used to simplify the comparison of training performance. However, not all algorithms use the same number of function evaluations per iteration with GOLS-I coupled to different algorithms. Each function evaluation consists of calling only the gradient, $\tilde{\boldsymbol{g}}(\boldsymbol{x})$.}
	\label{tbl_cifar_fvals}
\end{table}

In the case of GOLS-I SGD, GOLS-I Adagrad and GOLS-I Adam, the step sizes quickly stabilize around a particular order of magnitude during training. As observed in Section \ref{sec_golsAlgs}, this magnitude differs for every algorithm, yet is automatically recovered by GOLS-I. The step sizes remain relatively constant, with oscillations of up to half an order of magnitude. Consistent step sizes significantly reduce the total cost of training. As training continues, these oscillations can increase to 4 orders of magnitude. We postulate that this increasing trend occurs due to mini-batch samples that contain observations, that are far from the decision boundary, resulting in small gradients; and mini-batches that are close to the decision boundary, resulting in large gradients. As GOLS-I adapts, it randomly alternates between these cases causes increasing the step sizes for small gradients, and the converse for larger gradients. To compensating for this behaviour, \cite{Smith2017} propose gradually increasing the batch size during training.


\begin{table}[h!]
	\centering
	\scalebox{0.87}{
		\begin{tabular}{|c|c|c|c|c|}
			\hline 
			& \textbf{LS-LBFGS} & \textbf{LS-SGD} & \textbf{LS-Adagrad} & \textbf{LS-Adam, $\beta_1 = 0$} \\ 
			\hline 
			\textbf{Accuracy in \%} & 92.11 & 92.97 & 92.63 & 92.58 \\ 
			\hline 
		\end{tabular}
	} 
	\caption{Maximum achieved test classification accuracy after 40,000 training iterations on the CIFAR10 dataset with various training algorithms in combination with GOLS-I.}
	\label{tbl_cifar_maxAcc}
\end{table}

All algorithms considered in this investigation trained the model to a test classification accuracy of above 90\%, which is evident from the plateau in test classification, see Figure~\ref{fig_cifar}(d). Although the search directions vary in formulation for different algorithms, the outcome is comparable with the use of GOLS-I to determine step sizes. The maximum test classification accuracy achieved by each algorithm is shown in Table~\ref{tbl_cifar_maxAcc}. This study demonstrated that GOLS-I is also useful for more substantial, practical problems. Provided that no momentum terms are present in the algorithms, vastly different algorithms can have comparable training performance with the use of GOLS-I. This is in contrast to an investigation by \citet{Mukkamala2017}, which used {\it a priori} selected learning rates and learning rate schedules to solve the same problem. The authors' study showed a significant variance in training error between different learning rate schedules; and reported a best test classification accuracy of 86\% after the equivalent of $156,000$ function evaluations. An equivalent result was achieved in our investigation after $\pm 7,000$ function evaluations for GOLS-I SGD, GOLS-I Adagrad and GOLS-I Adam. Test classification accuracies achieved in our investigation are consistent with other studies \citep{He2016,Huang2018}, which implement manually tuned optimizers. \cite{He2016} make use of LS-SGDM with a $\alpha_{n,I_n}=0.1$, which is reduced by an order of magnitude at {\it a priori} determined intervals, for a total of 64,000 training iterations with a mini-batch size of $|\mathcal{B}_{n,i}|=128$, resulting in a test error of around 91\%. \cite{Huang2018} employ LS-Adam with $\alpha_{n,I_n}=0.01$, using a larger mini-batch size of $|\mathcal{B}_{n,i}|=512$ to achieve a test classification accuracy of 93.52\%. We achieve an average test classification error of 92.6\% over GOLS-I SGD, GOLS-I LBFGS, GOLS-I Adagrad and GOLS-I Adam ($\beta_1=0 $) with a mini-batch size of $|\mathcal{B}_{n,i}|=128$, confirming that GOLS-I determined the step sizes of selected algorithms effectively, without requiring any prior parameter tuning.

\section{Conclusion}

This study demonstrated that stochastic gradient-only line searches that are inexact (GOLS-I) allow for a generalized strategy to determine the learning rates of different training algorithms as opposed to conducting extensive hyper-parameter tuning studies. We considered 23 datasets with sizes varying between 150 and 70 000 observations, input dimensions ranging between 4 and 3072 and output dimensions between 2 and 29. Network architectures considered were feed-forward networks with 1 to 6 hidden layers of varying sizes and a convolutional architecture with residual connections. In addition, a total of seven training algorithms, namely: Gradient Descent, Gradient Descent with momentum, Nesterov's Accelerated Gradient Descent, Adagrad, Adadelta, Adam and LBFGS were considered. Our investigations have shown that algorithms with momentum terms do not perform well with GOLS-I. However, if it is possible to remove the momentum component from the algorithm, performance with GOLS-I drastically improves, which was demonstrated on both shallow and deep feed-forward neural networks. We also showed that without any parameter tuning, the performance of GOLS-I with LS-SGD, LS-BFGS, LS-Adagrad, and LS-Adam is competitive with other studies, which implement manually selected learning rate schedules. 


The use of GOLS-I avoids the need for manual parameter tuning or computationally expensive global parameter optimization approaches to determine an effective average fixed learning rate or explore learning rate schedule parameters. Gradient-only line search strategies on average require more than one gradient evaluation per iteration, making them more expensive than fixed-step size approaches per iteration. However, GOLS-I remains to our knowledge most effective adaptive learning rate optimization method to date. Additionally, our investigations have shown, that incorporating GOLS-I into the LBFGS algorithm shows promise for future improvements. This re-opens questions concerning the feasibility of conjugate gradient or Quasi-Newton methods in stochastic environments such as neural network training.

\acks{The Centre for Asset and Integrity Management (C-AIM) in the Department of Mechanical and Aeronautical Engineering, University of Pretoria, Pretoria, South Africa supported this work. We gratefully acknowledge the NVIDIA corporation, for supporting our research through the NVIDIA GPU grant.}


\newpage

\appendix


\section{Investigated optimization algorithms}
\label{sec_algs}

\subsection{Line Search Stochastic Gradient Descent (LS-SGD)}

Stochastic Gradient Descent (SGD) \citep{Robbins1951} is based on the steepest descent algorithm \citep{Arora2011}, but uses the dynamic MBSS loss function approximation, $\tilde{\boldsymbol{g}}(\boldsymbol{x})$ as its search directions. When a {\it a priori} learning rate schedule has been selected, SGD is equivalent to a subgradient approach \citep{Boyd2003}. We use line searches to determine its learning rates, called LS-SGD, in Algorithm~\ref{alg_LS-SGD} and coupled it with GOLS-I:


\begin{algorithm}[H]
	\DontPrintSemicolon 
	
	Set $n=0$ and choose the initial weights $\boldsymbol{x}_0$ \;
	
	\While{stop criterion not met}
	{
	Compute $\tilde{\boldsymbol{g}}(\boldsymbol{x}_n)$ \;
	Define the search direction, $\boldsymbol{d}_n = -\tilde{\boldsymbol{g}}(\boldsymbol{x}_n)$ \;
	Set the step length, $\alpha_{n,I_n}$, using a line search \;
	Update $\boldsymbol{x}_{n+1} = \boldsymbol{x}_n + \alpha_{n,I_n} \boldsymbol{x}_n$ \;
	}
	
	\caption{{\sc LS-SGD}: Line Search Stochastic Gradient Descent}
	\label{alg_LS-SGD}
\end{algorithm}

\subsection{Line Search Stochastic Gradient Descent with Momentum (LS-SGDM)}

The addition of a momentum term to the steepest descent formulation allows for a fraction of the previous update step to be added to the current step \citep{Rumelhart1988}. This emulates the behaviour of "momentum" in a physical system. The rationale behind this approach is to allow the algorithm to escape local minima with the aid of this "momentum". A consequence thereof is that ascent steps can be taken, in particular, if the momentum parameter is large. An outline of the method with a line search is given in Algorithm~\ref{alg_LS-SGDM}:


\begin{algorithm}[H]
	\DontPrintSemicolon 
	
	Set $n=0$ and choose initial weights $\boldsymbol{x}_0$, momentum constant $\gamma_m=0.9$, an initial update term $\boldsymbol{c}_{0} = \bar{\boldsymbol{0}}$ \;
	
	\While{stop criterion not met}
	{
	Compute $\tilde{\boldsymbol{g}}(\boldsymbol{x}_n)$ \;
	Define the descent direction $\boldsymbol{d}_n = -\tilde{\boldsymbol{g}}(\boldsymbol{x}_n)$ \;
	Set the step length, $\alpha_{n,I_n}$, using a line search\;
	Define the update step $\boldsymbol{c}_{n+1} = \alpha_{n,I_n} \boldsymbol{d}_n + \gamma_m \boldsymbol{c}_{n}$ \;
	Update $\boldsymbol{x}_{n+1} = \boldsymbol{x}_n + \boldsymbol{c}_n $ \;
	}
	
	\caption{{\sc LS-SGDM}: Line Search Stochastic Gradient Descent using Momentum}
	\label{alg_LS-SGDM}
\end{algorithm}

\subsection{Line Search Nesterov Accelerated Gradient Descent (LS-NAG)}

Nesterov's Accelerated Gradient Descent algorithm \citep{Nesterov1983} can be seen as an extension of the momentum strategy. In this case the gradient vector for the update step is evaluated only once the momentum term has been added to the current solution. This, therefore, results in a less naive implementation of the momentum concept. This method, as used with a line search, is given in Algorithm~\ref{alg_LS-NAG}:


\begin{algorithm}[H]
	\DontPrintSemicolon 
	
	Set $n=0$ and choose initial weights $\boldsymbol{x}_0$, an initial update term $\boldsymbol{c}_{0} = \bar{\boldsymbol{0}}$, momentum constant $\gamma_m = 0.5$ \;
	
	\While{stop criterion not met}
	{
	Compute $\tilde{\boldsymbol{g}}(\boldsymbol{x}_n + \gamma_m \boldsymbol{c}_{n})$ \;
	Define the search direction $\boldsymbol{d}_n = -\tilde{\boldsymbol{g}}(\boldsymbol{x}_n + \gamma_m \boldsymbol{c}_{n})$ \;
	Define the step length, $\alpha_{n,I_n}$, using a line search \;
	Define the update step $\boldsymbol{c}_{n+1} = \alpha_{n,I_n} \boldsymbol{d}_n + \gamma_m \boldsymbol{c}_{n}$ \;
	Update $\boldsymbol{x}_{n+1} = \boldsymbol{x}_n + \boldsymbol{c}_{n+1} $ \;
	}
	
	\caption{{\sc LS-NAG}: Line Search Nesterov Accelerated Gradient Descent}
	\label{alg_LS-NAG}
\end{algorithm}

\subsection{Line Search Adagrad}

The Adagrad algorithm \citep{Duchi2011} performs steepest descent updates with an integrated learning rate scheme independently on each weight. This means that a learning rate is assigned to every dimension separately. The learning rate magnitude is a function of the sum of the current, as well as previous squared gradient magnitudes. The learning rate schedule is constructed such that it biases higher learning rates for dimensions that have a flat slope (low partial derivative magnitudes), and assigns lower learning rates to dimensions with large slopes (high partial derivative magnitudes). We determine the learning rate for Adagrad using a line search in Algorithm~\ref{alg_LS-Adagrad}.


\begin{algorithm}[H]
	\DontPrintSemicolon 
	
	Set $n=0$, $ \boldsymbol{v}_{0} = \bar{\boldsymbol{0}}$, $ \boldsymbol{c}_{0} = \bar{\boldsymbol{0}}$ and choose the initial weights $\boldsymbol{x}_0$ \;
	
	\While{stop criterion not met}
	{
	Compute $\tilde{\boldsymbol{g}}(\boldsymbol{x}_n)$ \;
	Define $\boldsymbol{c}_n = -\tilde{\boldsymbol{g}}(\boldsymbol{x}_n)$ \;
	Calculate $ \boldsymbol{v}_{n+1} = \boldsymbol{c}_n \odot \boldsymbol{c}_{n} + \boldsymbol{v}_{n}$ 
	with $\odot$ indicating the element-wise multiplication or Hadamard product \citep{Reams1999}	\;
	Define the components of the search direction $ \boldsymbol{d}_n = (\boldsymbol{v}_{n+1}+\epsilon \bar{\boldsymbol{1}})^{\circ-\frac12} \odot \boldsymbol{c}_{n}$, with $\bar{\boldsymbol{1}}$ indicating a vector with all elements one and the superscript  $\circ$ the Hadamard power or the power of each element in the vector $(\boldsymbol{v}_{n+1}+\epsilon \bar{\boldsymbol{1}})$ to $-\frac12$. \;
	Set the step length, $\alpha_{n,I_n} $, using a line search \;
	Update $\boldsymbol{x}_{n+1} = \boldsymbol{x}_n + \alpha_{n,I_n} \boldsymbol{d}_n $ \;
	}
	
	\caption{{\sc LS-Adagrad}: Line Search Adagrad}
	\label{alg_LS-Adagrad}
\end{algorithm}

\subsection{Line Search Adadelta}
\label{sec_LS-Adadelta}

A disadvantage of Adagrad is that the accumulation of all the past gradients in the $\boldsymbol{v}_n$ term causes the $\boldsymbol{v}_n$ term to become large over time, diminishing the search direction.. This causes the overall learning rate to decrease and can cause slow progress in the later stages of training. Adadelta \citep{Zeiler2012} is an extension of Adagrad, which makes use of an exponentially decaying average for $\boldsymbol{v}_n$, such that a reasonable learning rate remains throughout training. It also implements an update magnitude rule in the form of an exponentially decaying average of the previous updates, $\boldsymbol{m}_n$. We add a line search to this method in Algorithm~\ref{alg_LS-Adadelta}:

%
%
%

\begin{algorithm}[H]
	\DontPrintSemicolon 
	
	Set $n=0$, $ \boldsymbol{v}_{0} = \bar{\boldsymbol{0}}$, $ \boldsymbol{d}_{0} = \bar{\boldsymbol{1}}$ and choose the initial weights $\boldsymbol{x}_0$ and  $\beta = 0.9$ \;
	
	\While{stop criterion not met}
	{
	Compute $\tilde{\boldsymbol{g}}(\boldsymbol{x}_n)$ \;
	Define $\boldsymbol{c}_n = -\tilde{\boldsymbol{g}}(\boldsymbol{x}_n)$ \;
	Calculate $ \boldsymbol{v}_{n+1} = (\beta - 1)\boldsymbol{c}_n \odot \boldsymbol{c}_{n} + \beta \boldsymbol{v}_{n}$ \;
	Calculate $ \boldsymbol{m}_{n+1} = (\beta - 1)\boldsymbol{d}_{n} \odot \boldsymbol{d}_{n} + \beta \boldsymbol{m}_{n}$ \;
	Define the components of the search direction 
	$ \boldsymbol{d}_{n+1} = (\boldsymbol{m}_{n+1}+\epsilon \bar{\boldsymbol{1}})^{\circ\frac12} \odot (\boldsymbol{v}_{n+1}+\epsilon \bar{\boldsymbol{1}})^{\circ-\frac12} \odot \boldsymbol{c}_{n}$ \;
	Determine $\alpha_{n,I_n}$, using a line search \;
	Define $\boldsymbol{x}_{n+1} = \boldsymbol{x}_n + \alpha_{n,I_n} \boldsymbol{d}_{n+1} $ \;
	}
	
	\caption{{\sc LS-Adadelta}: Line Search Adadelta}
	\label{alg_LS-Adadelta}
\end{algorithm}

\subsection{Line Search Adam}
\label{sec_LS-Adam}

Adaptive Moment Estimation (Adam) \citep{Kingma2015} also makes use of different learning rates for independent components of $\boldsymbol{x}_n$. These learning rates are a function of exponentially decaying past averages. In this case these are obtained from the first moment (the mean) $\boldsymbol{m}_n$ and the second moment (the centred variance) $\boldsymbol{v}_n$ of the past gradients. Due to the initial values for both these variables being chosen to be $\boldsymbol{0}$, the initial learning rates tend to be too small, resulting in slow training in the beginning. To account for this, the respective bias-corrected estimates are used, $\hat{\boldsymbol{m}}_n$ and $\hat{\boldsymbol{v}}_n$. We determine its learning rates using a line search in Algorithm~\ref{alg_LS-Adam}:

%

\begin{algorithm}[H]
	\DontPrintSemicolon 
	
	Set $n=0$, $ \boldsymbol{m}_{0} = \bar{\boldsymbol{0}}$, $ \boldsymbol{v}_{0} = \bar{\boldsymbol{0}}$, and choose the initial weights $\boldsymbol{x}_0$,  $\beta_1= 0.9$ ($\beta_1= 0$) and $\beta_2 = 0.999$ \;
	
	\While{stop criterion not met}
	{
		Compute $\tilde{\boldsymbol{g}}(\boldsymbol{x}_n)$ \;
		Define $\boldsymbol{c}_n = \tilde{\boldsymbol{g}}(\boldsymbol{x}_n) $ \;
		Define $\boldsymbol{m}_{n+1} = \beta_1 \boldsymbol{m}_{n} + (1-\beta_1) (\boldsymbol{c}_{n})$ \;
		Define $\boldsymbol{v}_{n+1} = \beta_2 \boldsymbol{v}_{n} + (1-\beta_2) (\boldsymbol{c}_{n} \odot \boldsymbol{c}_{n})$ \;
		Calculate $\hat{\boldsymbol{m}}_{n+1} = \frac{\boldsymbol{m}_{n+1}}{1-\beta_1}$ \;
		Calculate $\hat{\boldsymbol{v}}_{n+1} = \frac{\boldsymbol{v}_{n+1}}{1-\beta_2}$ \;
		Define the components of the search direction $ \boldsymbol{d}_n = (\boldsymbol{\hat{v}}_{n+1}^{\circ\frac12}+ \epsilon \bar{\boldsymbol{1}})^{\circ -1} \odot \hat{\boldsymbol{m}}_{n+1} $ \;
		Set the step length, $\alpha_{n,I_n}$, using a line search \;
		Update $\boldsymbol{x}_{n+1} = \boldsymbol{x}_n + \alpha_{n,I_n} \boldsymbol{d}_n$ \;
	}
	
	\caption{{\sc LS-Adam}: Line Search Adam}
	\label{alg_LS-Adam}
\end{algorithm}

\section{Inexact Line Search: Gradient-Only Line Search that is Inexact (GOLS-I)}
\label{sec_golsi}


Parameters used for this method are: $\eta = 2$, $c_2 = 0.9$, $\alpha_{min} = 10^{-8}$ and $\alpha_{max} = 10^7$.  $F'_n(\alpha) =  \boldsymbol{\tilde{g}}(\boldsymbol{x}_n + \alpha \cdot \boldsymbol{d_n})\boldsymbol{d_n} $.

\begin{algorithm}[H]
	\DontPrintSemicolon 
	\KwIn{$F_n'(\alpha)$, $\boldsymbol{d}_n$ , $\alpha_{n,0}$}
	\KwOut{$\alpha_{n,I_n}$, $k$}
	
	Define constants: $\alpha_{min}=10^{-8}$, flag = 1,  $k = 0$, $\eta = 2$, $c_2=0.9$ \;
	$\alpha_{max} = min(\frac{1}{||\boldsymbol{d}_n ||_2}, 10^7)$ \;
	
	Evaluate $ F_n'(0) $, increment $k$ (or use $ F_{n-1}'(\alpha_{n-1,I_n})$ without incrementing $k$)\;
	\If{ $\alpha_{n,0} > \alpha_{max}$ }{
		$ \alpha_{n,0} = \alpha_{max} $ 
	}
	\If{ $\alpha_{n,0} < \alpha_{min}$}{
		$ \alpha_{n,0} = \alpha_{min} $ 
	}
	Evaluate $ F_n'(\alpha_{n,0})$, increment $k$ \;
	Define $tol_{dd} = |c_2 F_n'(0)|$ \;
	
	\If{ $F_n'(\alpha_{n,0}) > 0$ and $\alpha_{n,0} < \alpha_{max}$ }{
		flag = 1, decrease step size 
	}
	\If{ $F_n'(\alpha_{n,0}) < 0$ and $\alpha_{n,0} > \alpha_{min}$ }{
		flag = 2, increase step size  
	}
	\If{  $F_n'(\alpha_{n,0}) > 0$ and $F_n'(\alpha_{n,0}) < tol_{dd}$ }{
		flag = 0, immediate accept condition
	}
	
	\While{flag $>0$}{ 
		\If {flag = 2}{
			$\alpha_{n,i+1} = \alpha_{n,i} \cdot \eta$ \;
			Evaluate $ F_n'(\alpha_{n,i+1}) $, increment $k$ \;
			
			\If{ $F_n'(\alpha_{n,i+1}) \geq 0 $}{
				flag = 0 \;
			}
			\If{ $\alpha_{n,i+1} > \frac{\alpha_{max}}{\eta}$}{
				flag = 0
			}	
		}
		\If {flag = 1}{
			$\alpha_{n,i+1} = \frac{\alpha_{n,i}}{\eta} $ \;
			Evaluate $ F_n'(\alpha_{n,i+1}) $, increment $k$ \;
			\If{$F_n'(\alpha_{n,i+1}) < 0$}{
				flag = 0
			}
			\If{ $\alpha_{n,i+1} < \alpha_{min} \cdot \eta$}{
				flag = 0
			}
		}
	}
	
	$\alpha_{n,I_n}$ = $\alpha_{n,i+1}$
	
	\caption{{\sc GOLS-I}: Gradient-Only Line Search that is Inexact \citep{Kafka2019jogo}}
	\label{alg_GOLS-I}
\end{algorithm}

\section{Heuristics for determining the number of hidden units}
\label{app_heuristics}

The number of hidden units in a single hidden layer feed-forward network, $H$, was chosen to be the smaller of two heuristics, $H_1$ and $H_2$ given as:
\begin{equation}
H_1 = \frac{\frac{M}{C_r} - K}{D+K+1},
\label{eq_nHN1}
\end{equation}
and 
\begin{equation}
H_2 = D-1.
\label{eq_nHN2}
\end{equation}

And therefore,

\begin{equation}
H = min(H_1,H_2),
\label{eq_nHNs}
\end{equation}

where $D$ is the number of input features of a given dataset, $M$ is the total number of observations, $C_r$ is a regression constant, and $H_1$ is rounded down to the nearest integer value.
The regression constant $C_r$ determines how rigid the model is, with $C_r > 1$ resulting in fewer parameters relative to the degrees of freedom in the data, and $C_r < 1$ resulting in more parameters than data points in the training dataset. In our investigation, this constant was set to $C_r = 1.5 $, which ensures that the model regresses through the data. Arguably, pruning \citep{Reed1993} is an alternative approach, but our heuristic approach is sufficient for this study with the benefit of simplicity.

\vskip 0.2in
\bibliography{BibC2}

\end{document}